\documentclass{article}

\usepackage[preprint]{ail_at_hku}

\usepackage[utf8]{inputenc}
\usepackage[T1]{fontenc}
\usepackage{hyperref}
\usepackage{url}
\usepackage{booktabs}
\usepackage{amsfonts}
\usepackage{nicefrac}
\usepackage{microtype}
\usepackage{xcolor}
\usepackage{graphicx}
\usepackage{multirow}
\usepackage{subcaption}
\usepackage{wrapfig}
\usepackage{longtable}
\usepackage{array}
\usepackage{ragged2e}
\usepackage{tabularx}
\usepackage{amsmath}
\usepackage{pdflscape}
\usepackage{amssymb}
\usepackage{pifont}

\title{Offline Reinforcement Learning for Plasma Control in Nuclear Fusion: Codebase and Benchmark}

\author{%
  \begin{minipage}{\linewidth}
    \centering
    \normalsize\sffamily\bfseries
    Yang Fu\textsuperscript{1,$\dagger$}
    \quad
    Haomin Bao\textsuperscript{2,$\dagger$}
    \quad
    Rohit Sonker\textsuperscript{3}
    \quad
    Xiaoyan Hu\textsuperscript{3}
    \\[3pt]
    Aravind Venugopal\textsuperscript{3}
    \quad
    Jeff Schneider\textsuperscript{3}
    \quad
    Jiayu Chen\textsuperscript{4,$\ddagger$}
    \\[8pt]
    \normalsize\rmfamily\normalfont
    \textsuperscript{1}Central South University
    \quad
    \textsuperscript{2}Chongqing University
    \\[3pt]
    \textsuperscript{3}Carnegie Mellon University
    \quad
    \textsuperscript{4}The University of Hong Kong
    \\[8pt]
    \small\rmfamily\normalfont
    \textsuperscript{$\dagger$}Equal contribution.
    \quad
    \textsuperscript{$\ddagger$}Corresponding author.
  \end{minipage}
}

\begin{document}

\maketitle

\begin{abstract}{https://github.com/LucasCJYSDL/Offline-RL-Kit-for-Nuclear-Fusion}{}{Jiayu Chen (jiayuc@hku.hk)}{}
  Offline reinforcement learning (RL) offers a promising route for developing plasma controllers from historical tokamak data, since online trial-and-error on real devices is costly and risky. However, progress in this direction remains difficult to measure due to the lack of a standardized offline RL benchmark for realistic multi-actuator, long-horizon plasma control problems in nuclear fusion. We introduce RL4F, an Offline Reinforcement Learning Benchmark for Plasma Control in Nuclear Fusion, providing closed-loop evaluation environments and baseline comparisons across four full-profile tracking tasks: rotation, density, temperature, and pressure. The dynamics function underlying the evaluation environment is built from historical discharge data from DIII-D, a real-world Tokamak. 
We evaluate a broad set of imitation learning and offline RL baselines under a unified protocol. We find that offline model-based RL methods obtain the best average performance on most objectives, although no single method dominates all tasks, highlighting the importance of dynamics modeling in complex, long-horizon plasma control tasks. To foster further research, we open-source the codebase, datasets, and evaluation framework, providing a benchmark not only for the fusion community but also for algorithm development in offline RL.

\end{abstract}

\section{Introduction}




Nuclear fusion offers a potential route toward abundant, low-carbon energy by harnessing the reactions that power stars~\citep{Gi2020}. The tokamak is one of the most promising confinement devices for achieving controllable nuclear fusion, but its operation requires real-time control of hot, unstable, and strongly coupled plasmas. Recent work has shown that reinforcement learning (RL) can be used to train such controllers. In particular, \citet{Degrave2022} demonstrated deep RL-based magnetic control on the TCV tokamak, and subsequent studies have extended RL-based plasma control toward tearing-mode avoidance, profile tracking, and ramp-down control~\citep{tracey2024towards,Seo2024,pmlr-v211-char23a,Wang2025}. These advances suggest that RL can complement conventional plasma-control design by directly optimizing feedback policies in high-dimensional tokamak control problems.

Developing RL controllers directly on real tokamaks is difficult to scale: tokamak operation is expensive, time-limited, and safety-critical. A natural way is to train candidate policies and evaluate their closed-loop behavior in simulation before possible deployment. Physics-based simulators, such as RAPTOR~\citep{felici2011realtime_raptor}, Forward Grad-Shafranov
Static~(FGE) simulator~\citep{carpanese2021}, and the more recent TORAX~\citep{torax2024arxiv}, provide reliable tools for forward modeling, trajectory optimization, and controller development. Previous work has demonstrated promising control performance for RL policies trained on these simulators~\citep{Degrave2022, tracey2024towards}. However, physics-based simulators are significantly more computationally expensive than traditional RL simulators, especially when RL agents explore the actuator space stochastically during training, which often results in slower convergence for the iterative solver. Moreover, adapting these simulators to a specific tokamak can require substantial modeling choices, parameter identification, and calibration.
An alternative approach is to learn control-oriented dynamics models from historical experimental data, yielding simulation environments tied more directly to a particular device~\citep{pmlr-v211-char23a,sonker2026offline}. Despite recent progress, RL-based profile control remains challenging: plasma profiles are high-dimensional spatial quantities, their dynamics are nonlinear and coupled across multiple actuators, and practical RL controllers still face difficulties in reward specification, steady-state tracking bias, and sample efficiency~\citep{tracey2024towards}. These challenges motivate a standardized benchmark for data-driven training and offline evaluation of RL algorithms for plasma control in tokamaks.

Our primary contribution is RL4F, a unified benchmark for offline RL in tokamak profile control. A profile refers to the radial spatial variation of a plasma quantity from the core, i.e., the innermost region of the plasma, to the edge, i.e., the outermost region. It is characterized by four key features. 1) \textbf{Realistic pre-deployment workflow.} We first train a reference dynamics model from historical DIII-D experimental discharges, and then use this model to generate trajectories for offline policy learning. Candidate algorithms learn only from the model-generated datasets and are evaluated in closed loop on the reference dynamics model, mirroring the practical setting in which controllers must be developed from past experimental data before any real-machine test. 
2) \textbf{Multi-task profile tracking.} The benchmark covers four full-profile tracking tasks, including rotation, density, temperature, and pressure, which expose different control difficulties and plasma response channels. 3) \textbf{Scenario-relevant actuator space.} Policies act through a shared action space consisting of neutral-beam power, neutral-beam torque, gas puffing, and electron-cyclotron heating, covering the main heating, momentum-input, and fueling channels relevant to profile control for DIII-D, a tokamak device located in San Diego. 4) \textbf{Large-scale data.} The benchmark contains 5,882 shots and 945,828 transitions after filtering, with fixed training, validation, and test splits. To our knowledge, this is the first benchmark specifically designed for offline RL in fusion plasma control. 

We evaluate representative baselines spanning imitation learning, model-free offline RL, and model-based offline RL under the same closed-loop protocol and profile-level tracking metrics.
Our evaluation shows that model-based methods generally outperform model-free baselines, while no single algorithm dominates across all profile objectives, highlighting tokamak profile control as a challenging offline RL benchmark.


\section{Related Work}

\textbf{Fusion Plasma Control.}\quad 
Nuclear fusion is a leading candidate for sustainable power generation. A central challenge is the control of plasma profiles to achieve stable, high-performance operation. Conventional plasma control typically relies on precomputed feedforward coil current trajectories~\citep{Walker01112006} together with feedback loops for individual target quantities. Such profile control systems have been implemented on several tokamaks, including JET for safety \textit{q}-profile control~\citep{DMoreau_2003}, TCV for \textit{q}-profile and electron-temperature control~\citep{7171844}, and EAST for \textit{q}-profile control~\citep{9658937}. These approaches have demonstrated promising performance in a wide range of discharges, but design can be challenging and time-consuming, especially in plasma scenarios where the control quantities are high-dimensional or strongly coupled. More recently, reinforcement learning (RL) has emerged as a new framework for designing feedback controllers in fusion systems. \citet{Degrave2022} train an RL-based controller under the Maximum-a-Posteriori~(MPO)~\cite{abdolmaleki2018maximum} framework to track the location, current, and shape across a diverse set of plasma configurations. This line of work is further extended by~\citet{tracey2024towards}. \citet{pmlr-v211-char23a} develop an offline RL framework for tracking $\beta_N$ and plasma rotation quantities and train the controller using Proximal Policy Optimization~(PPO), which is later utilized to design a Bayesian optimization~(BO)-style controller~\citep{sonker2025multitimescale} for mitigating tearing instabilities. \citet{Seo2024} apply the Deep Deterministic Policy Gradient~(DDPG)~\cite{lillicrap2019} approach to maintain high-pressure plasma at the H-mode performance while avoiding tearing instabilities. \citet{Wang2025} trains RL policies to avoid disruptions during the ramp-down phase. 

\textbf{Benchmarks for Offline RL.}\quad
The well-known D4RL dataset~\citep{fu2021} benchmarks offline RL in challenging robotic control scenarios with biased data distributions. The Dope benchmark~\citep{fu2021benchmarksdeepoffpolicyevaluation}, which builds upon D4RL and RL Unplugged \citep{NEURIPS2020_51200d29}, focuses on off-policy evaluation. \citet{NEURIPS2022_9cd828eb} propose NeoRL to mitigate the gap between earlier offline RL benchmarks and real-world scenarios. \citet{liu2023datasetsbenchmarksofflinesafe} present a benchmarking suite that facilitates the development and evaluation of offline safe RL algorithms in both the training and deployment phases. \citet{park2025ogbench} propose OGBench for offline goal-conditioned RL. Compared with these existing benchmarks, fusion plasma control tasks present unique challenges, including highly nonlinear and stochastic dynamics, partial observability, and safety-critical operating constraints. To the best of our knowledge, RL4F is the first offline RL benchmark tailored to fusion plasma control, a critical real-world operational scenario.

\section{Simulator}

\textbf{Problem Setup.} We formulate tokamak profile control as a finite-horizon Markov decision process (MDP) $\mathcal{M}=(\mathcal{S},\mathcal{A},P,r,\rho_0,\gamma,H)$, where
$\mathcal{S}$ is the state space, $\mathcal{A}$ is the action space,
$P(s'|s,a)$ is the transition distribution, $r(s,a)$ is the reward function, $\rho_0$ is the initial-state distribution, $\gamma$ is the discount factor, and $H$ is the episode horizon. A policy $\pi(a|s)$ aims to maximize the expected discounted return
\begin{equation}
    J(\pi)=
    \mathbb{E}_{\pi,P,\rho_0}
    \left[
    \sum_{t=0}^{H-1}\gamma^t r(s_t,a_t)
    \right].
\end{equation}
\begin{figure}[t]
    \centering
    \includegraphics[width=\linewidth]{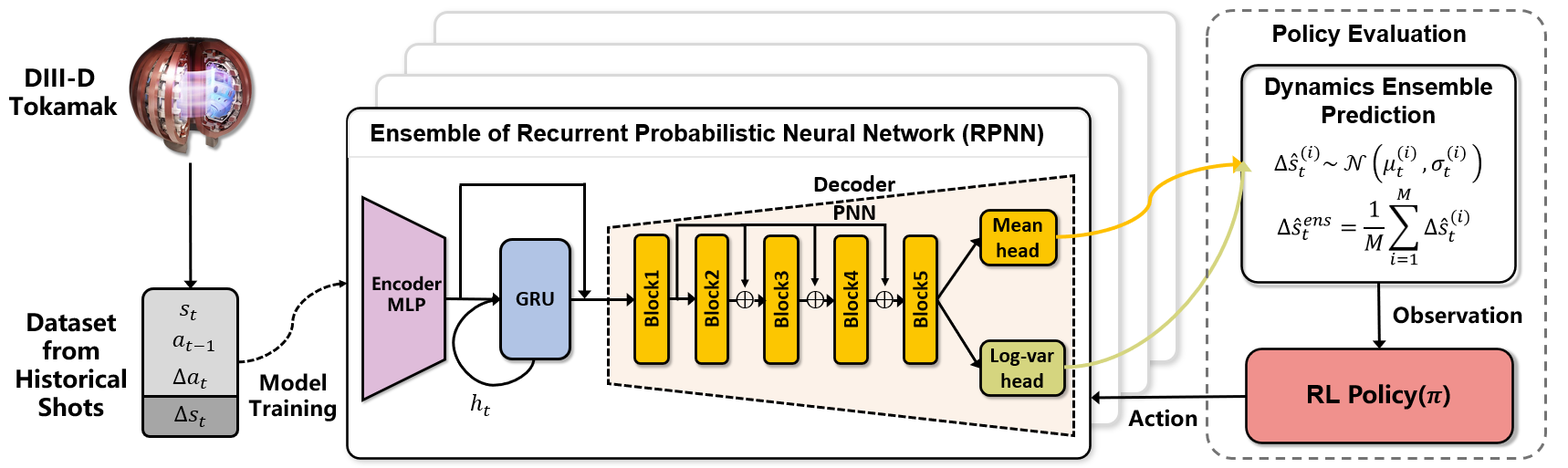}
    \vspace{-.05in}
\caption{
The reference RPNN dynamics ensemble is trained from historical DIII-D operational data and used as the closed-loop environment for evaluating trained policies.}

    \label{fig:dynamic1}
    \vspace{-.1in}
\end{figure}
In offline reinforcement learning, the agent cannot interact with the real environment during training~\citep{levine2020offline} and is instead given a fixed dataset of trajectories $\mathcal{D}=\{\tau_i\}_{i=1}^N$, where each trajectory consists of transitions $(s_t,a_t,r_t,s_{t+1})$ collected by an unknown behavior policy. This setting is well matched to tokamak control, where online trial-and-error is expensive and risky.

We use a recurrent probabilistic neural network (RPNN) to learn plasma dynamics from trajectory data, following prior data-driven tokamak dynamics modeling and offline model-based control work~\citep{pmlr-v211-char23a,sonker2026offline}. Given the current plasma state $s_t$, the previous actuator setting $a_{t-1}$, and the actuator increment  $\Delta a_t = a_t - a_{t-1}$, the model predicts a distribution over the next state change, $\Delta s_t = s_{t+1} - s_t$. Specifically, the RPNN outputs the parameters of a Gaussian distribution, $\mu_t, \log \sigma_t^2 = f_\theta(s_t, a_{t-1}, \Delta a_t)$, and the next state is advanced autoregressively by adding the predicted state change to the current state. This recurrent probabilistic formulation allows the model to capture history-dependent plasma evolution while providing uncertainty estimates for long-horizon rollouts. A schematic of the dynamics-model training 
workflow is shown in Figure~\ref{fig:dynamic1}.

\begin{wrapfigure}{r}{0.35\textwidth}
    \centering
    \vspace{-.15in}
    \includegraphics[width=0.95\linewidth]{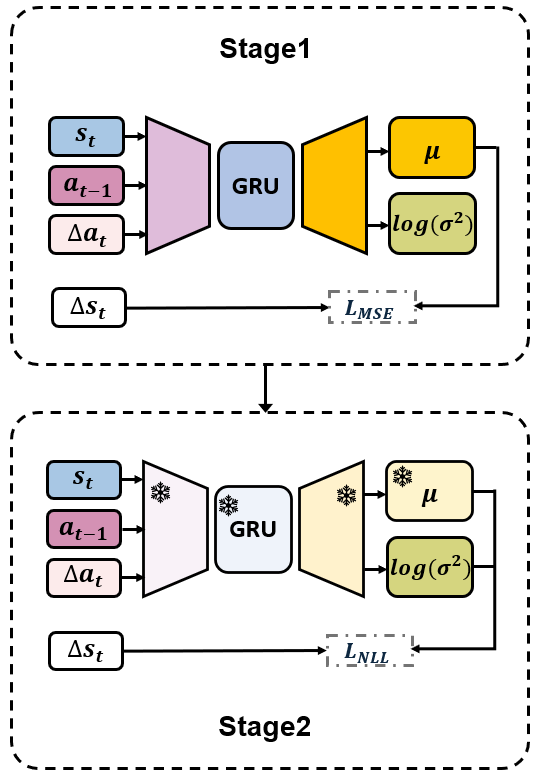}
    \vspace{-.1in}
    \caption{Two-stage training procedure for the dynamics model.}
    \label{fig:two_stage_dynamics_training}
\end{wrapfigure}

\textbf{Dynamics Modeling.} Given the limited accessibility of real tokamak devices, we train a reference dynamics model as a digital twin. The reference dynamics model is trained from historical DIII-D experimental discharges and is used to generate offline training data and to provide the closed-loop evaluation environment. As shown in Figure~\ref{fig:two_stage_dynamics_training},
we adopt a two-stage RPNN training procedure \citep{sonker2026offline}.

The first stage optimizes the predictive mean using a mean-squared error objective. The second stage initializes from the first-stage checkpoint, freezes the predictive backbone, and trains the log-variance head using a negative log-likelihood objective, thereby calibrating uncertainty without changing the learned mean dynamics. Following prior model-based control practice \citep{chen2025policy}, we train a bootstrapped ensemble of RPNNs; the predicted log variance captures aleatoric uncertainty, while disagreement across ensemble members provides an estimate of epistemic uncertainty. More training details are given in Appendix~\ref{app:rpnn}. 



We train a 25-member RPNN ensemble on roughly 18,000 historical DIII-D experimental discharges, spanning nearly a decade of data collection. Each shot contains approximately four seconds of data sampled at 20~ms.
For profile quantities, including electron temperature, ion temperature, density, pressure, rotation, and the safety-factor $q$ profile, we use ZipFIT reconstructions~\citep{logan2018omfit}, which provide smooth, physics-constrained profile estimates. Following prior tokamak dynamics modeling work~\citep{pmlr-v211-char23a}, we reduce the dimension of profile quantities with PCA before dynamics-model training. The held-out predictive fidelity of the trained ensemble is reported in Appendix~\ref{app:rpnntrain}.
To simulate a tokamak experiment, i.e., a ``shot'', we use the ensemble of dynamics models $\{f_{\theta_i}\}_{i=1}^{25}$. Each ensemble member represents a plausible version of the device dynamics, and the ensemble captures
uncertainty in the true transition model. Notably, RL policies trained on the same ensemble of dynamics models have been verified to be effective for profile control on DIII-D~\citep{sonker2026offline}.

\textbf{Synthetic Dataset for Offline RL.} The offline dataset is generated by initializing rollouts from real flat-top shot states after a warm-up period, replaying the real actuator sequence, and autoregressively rolling out the reference dynamics model. 
At timestep $t$, the measured next state is replaced by the ensemble prediction
\begin{equation}
    \tilde{\Delta s}_t^{(m)}
    \sim
    \mathcal{N}\!\bigl(
        \mu_t^{(m)},
        \operatorname{diag}((\sigma_t^{(m)})^2)
    \bigr),
    \qquad
    \hat{s}_{t+1}
    =
    \hat{s}_{t}
    +
    \frac{1}{M}
    \sum_{m=1}^{M}
    \tilde{\Delta s}_t^{(m)},
    \quad M=25,
\end{equation}
where $\mu_t^{(m)}$ and $\sigma_t^{(m)}$ denote the mean and standard deviation predicted by the $m$-th RPNN ensemble member at time step $t$, and $\tilde{\Delta s}_t^{(m)}$ denotes a sampled state increment drawn from the corresponding Gaussian predictive distribution. The generated state $\hat{s}_{t+1}$ is then fed back as the input for the next timestep until the end of the shot segment. This procedure preserves the actuator schedules and shot segmentation of the real experiments, while replacing the measured plasma evolution with model-generated dynamics. 

We restrict the synthesized data to the flat-top phase of each discharge and remove outlier values, resulting in 5,882 shots and 945,828 timesteps. From these filtered shots, we randomly sample 300 shots for validation and 300 shots for testing, and use the remaining 5,282 shots for training. This yields 849,977 training timesteps, 48,010 validation timesteps, and 47,841 test timesteps, with a maximum shot length of 180 timesteps after filtering. Offline RL and imitation learning algorithms are trained only on the generated training dataset and evaluated on the reference dynamics model.

\section{Benchmarking Tasks}




We benchmark algorithms on full-profile tracking tasks in tokamak plasma control. Unlike scalar tracking, profile tracking requires regulating high-dimensional spatial quantities across the normalized radial coordinate. This is challenging for offline RL because plasma profiles evolve over long horizons under nonlinear, stochastic dynamics, while policies must infer corrective actions from a fixed dataset.


In this work we address profile tracking across four plasma quantities - rotation, density, electron temperature, and pressure. \textbf{Rotation profile control} is beneficial as differential rotation control impacts suppression of tearing instabilities \citep{richner2024use}. Moreover, edge rotation affects penetration of neutral gas, leading to asymmetric fueling \citep{emdee2024influence, wilkie2024reconstruction}, a process crucial for modern high-performance fusion scenarios because it helps manage plasma exhaust and maintain the high density required for fusion without damaging the reactor walls. \textbf{Density profile control} regulates the core fuel inventory and fusion reaction rate while avoiding excessive edge density, radiation cooling, confinement degradation, and density-limit disruptions. Experiments and density-limit studies show that peaked or optimized density profiles can extend the operational range and support high-density, high-confinement tokamak plasmas \citep{greenwald2002density}. \textbf{Electron temperature profile control} is important because the profile gradient directly affects heat transport and regulation of this profile helps maintain core thermal energy and confinement while keeping the plasma within favorable regimes \citep{PhysRevResearch.7.L012004}. Finally, \textbf{Pressure profile control} is important because the achievable normalized pressure and MHD stability limit depend strongly on the shape of the pressure and current-density profiles. A poorly placed pressure gradient can drive instabilities, while an optimized pressure profile helps sustain high stored energy without crossing stability limits \citep{strait1994stability}. The profile tracking tasks can be modeled as MDPs. We define their observation space, action space, and reward function as follows, the effectiveness of which has been testified in real-world fusion control experiments~\citep{sonker2026offline}.

\textbf{Observation Space.} We include the present profile, the current target profile, and a future target profile shifted 10 timesteps ahead. Following the PID-inspired observation design used in practical tokamak RL control~\citep{tracey2024towards}, we also include proportional error terms (P-terms) for both the current and future targets. These terms are defined as the difference between the target profile and the present profile, and provide the policy with explicit feedback on the current tracking error.

\textbf{Target Sampling.} Randomly set targets may be infeasible within a given regime. To ensure feasibility, we construct targets from reference trajectories by sampling two time points and forming a step function with three segments (first–second–first).  This strategy ensures that all generated targets are consistent with the system dynamics while introducing sufficient variation to enrich the training distribution.

\textbf{Action Space.} The policy controls NBI power, NBI torque, gas-puffing voltage, and total ECH power. NBI torque directly influences the plasma rotation profile at the beam deposition location, which is typically situated near the plasma core. Both torque and power originate from the same neutral beam source, playing a central role in sustaining the plasma. We follow a counter-beam configuration of neutral beams, which allows independent control of injected power and torque.

\textbf{Reward Function.} The reward function is the mean squared tracking error on the whole profile (33 dimensions per timestep), which is reconstructed from corresponding PCA components.
\begin{equation}
r_t = -\left\| p_{\mathrm{target}}(t) - p(t) \right\|_2^2
\end{equation}
where $p_{\mathrm{target}}(t)$ is the target profile at time $t$, and $p(t)$ is the actual profile.

\section{Benchmarking Results}




%
\label{sec:benchmark}

We benchmark diverse baselines on four profile-tracking tasks. All policies are trained on a static dataset synthesized by the reference dynamics model and evaluated in closed loop on the same reference model, as illustrated in Figure~\ref{fig:dynamic1}. Following \citet{sonker2026offline}, evaluation rollouts use the mean prediction of the reference ensemble as the next-state estimate, and performance is measured by profile-tracking error on held-out test shots.

\paragraph{Baselines}
The evaluated methods include model-free offline RL baselines --TD3+Behavior Cloning (TD3BC) \citep{fujimoto2021td3bc}, Conservative Q-Learning (CQL) \citep{kumar2020cql}, 
Implicit Q-Learning (IQL) \citep{kostrikov2022iql}, Ensemble-Diversified Actor-Critic (EDAC) \citep{an2021edac}, and Mildly Conservative Q-Learning (MCQ) \citep{lyu2022mcq}; model-based offline RL baselines -- PPO \citep{schulman2017ppo}, Conservative Offline Model-Based Policy Optimization (COMBO) \citep{yu2021combo}, Model-based Offline Policy Optimization (MOPO) \citep{yu2020mopo}, Model-Bellman Inconsistency (MOBILE) \citep{sun2023mbi}, Robust Adversarial Model-Based Offline Reinforcement Learning (RAMBO) \citep{rigter2022rambo}, and Bayes Adaptive Monte Carlo Tree Search (BAMCTS) \citep{chen2026bamcts}; and Goal-Conditioned Imitation Learning (GCIL) \citep{ding2019gcil}. Full experiment details in Appendix~\ref{app:implementationdetails}

Model-based offline RL baselines train their own dynamics models from the synthetic trajectories generated by the reference model. We use the same RPNN ensemble architecture and two-stage training pipeline described in Section~3, again training a 25-member ensemble. The held-out predictive fidelity of the learned dynamics models (for offline RL) is summarized in Appendix~\ref{app:rpnntrain}. During model-based RL training, for each rollout, we sample one ensemble member and then generate the trajectory autoregressively by sampling from that member's predictive distribution, parameterized by its mean and log-variance. A new ensemble member is selected whenever a new rollout trajectory is sampled. Following \citet{chua2018deepreinforcementlearninghandful}, this procedure trains policies to perform across the range of plausible learned dynamics. 

\paragraph{Evaluation Metrics}
We evaluate the baselines on each profile tracking task using closed-loop simulation over 300 held-out test shots, with 10 random seeds per shot. At each timestep, the environment provides the target profile and the agent-achieved profile. 

We compute the tracking error at each normalized radial location 
$\psi_N$, where $N$ ranges uniformly from 0 (core) to 1 (edge). There are 33 radial locations in total. The per-location tracking metric is RMSE, computed from the squared deviation between the target and achieved profiles over the rollout horizon. We report the mean RMSE and its standard error (SE) across all rollout instances at six selected radial locations. Lower values indicate better tracking performance.

\begin{equation}
\overline{\mathrm{RMSE}}(\psi_N)
=
\frac{1}{NS}
\sum_{i=1}^{N}
\sum_{j=1}^{S}
\mathrm{RMSE}^{(i,j)}(\psi_N),
\end{equation}
\noindent where $N$ is the number of test shots and $S$ is the number of random seeds per shot. 
$\mathrm{RMSE}^{(i,j)}(\psi_N)$ denotes the per-rollout RMSE at radial location $\psi_N$ 
for shot $i$ and seed $j$.

\paragraph{Main Results}
Tables~\ref{tab:rotation}--\ref{tab:pressure} summarize the tracking performance of all baselines across the four tasks. Across all tasks, offline RL methods achieve substantially lower tracking errors than the goal-conditioned imitation learning baseline, indicating that purely imitation-based learning is insufficient for these long-horizon profile-tracking problems.

For rotation tracking, RAMBO achieves the lowest average RMSE (8.03), narrowly outperforming COMBO and MOPO. Model-free baselines perform substantially worse. Rotation profiles have pronounced radial variation from the plasma core to the edge and respond primarily through torque actuation, requiring sustained correction over long horizons after each target switch. In this setting, conservative or behavior-regularized model-free methods are more restricted in their ability to deviate from the behavior distribution within the static training dataset when active correction is needed.
\begin{table}[htbp]
    \centering
    \caption{Rotation profile tracking error (RMSE $\downarrow$). Values are reported as mean $\pm$ standard error over test shots.}
    \label{tab:rotation}

    \resizebox{\textwidth}{!}{%
        \begin{tabular}{lcccccccccccccc}
            \toprule
            \multirow{2}{*}{\textbf{Algorithm}} & \multicolumn{2}{c}{$\psi_{0.09}$} & \multicolumn{2}{c}{$\psi_{0.18}$} & \multicolumn{2}{c}{$\psi_{0.39}$} & \multicolumn{2}{c}{$\psi_{0.58}$} & \multicolumn{2}{c}{$\psi_{0.79}$} & \multicolumn{2}{c}{$\psi_{0.88}$} & \multicolumn{2}{c}{Average} \\
            \cmidrule(lr){2-3} \cmidrule(lr){4-5} \cmidrule(lr){6-7} \cmidrule(lr){8-9} \cmidrule(lr){10-11} \cmidrule(lr){12-13} \cmidrule(lr){14-15}
             & Mean & SE & Mean & SE & Mean & SE & Mean & SE & Mean & SE & Mean & SE & Mean & SE \\
            \midrule

            GCIL
            & 32.45 & 0.376 & 28.46 & 0.336 & 20.73 & 0.240 & 15.72 & 0.180 & 11.76 & 0.135 & 10.28 & 0.116 & 17.76 & 0.047 \\

            \midrule

            TD3BC
            & 25.10 & 0.318 & 20.99 & 0.274 & 14.81 & 0.189 & 11.35 & 0.141 & 8.69 & 0.104 & 7.79 & 0.092 & 13.13 & 0.038 \\

            CQL
            & 24.88 & 0.364 & 21.09 & 0.319 & 15.48 & 0.241 & 12.13 & 0.192 & 9.28 & 0.143 & 8.23 & 0.120 & 13.58 & 0.044 \\

            IQL
            & 29.64 & 0.439 & 25.72 & 0.391 & 18.85 & 0.275 & 14.47 & 0.203 & 10.87 & 0.149 & 9.48 & 0.129 & 16.23 & 0.051 \\

            EDAC
            & 26.30 & 0.427 & 20.78 & 0.359 & 14.43 & 0.261 & 11.41 & 0.210 & 9.10 & 0.157 & 8.31 & 0.129 & 13.30 & 0.048 \\

            MCQ
            & 28.45 & 0.344 & 24.08 & 0.310 & 17.19 & 0.224 & 13.22 & 0.167 & 10.10 & 0.120 & 8.90 & 0.102 & 15.10 & 0.043 \\

            \midrule

            PPO
            & 13.96 & 0.159 & 10.41 & 0.138 & 9.81 & 0.120 & 9.85 & 0.124 & 8.51 & 0.112 & 7.79 & 0.103 & 9.36 & 0.023 \\

            COMBO
            & 17.67 & 0.309 & 12.85 & 0.252 & 8.02 & 0.160 & 6.29 & 0.118 & 5.29 & 0.085 & 5.43 & 0.074 & 8.09 & 0.032 \\

            MOPO
            & 18.21 & 0.241 & 13.47 & 0.213 & 9.19 & 0.163 & 7.32 & 0.130 & 5.62 & 0.093 & 5.20 & 0.076 & 8.66 & 0.030 \\

            MOBILE
            & 48.07 & 0.729 & 41.59 & 0.620 & 30.98 & 0.438 & 24.50 & 0.338 & 18.93 & 0.258 & 16.37 & 0.221 & 26.86 & 0.083 \\
            
            RAMBO
            & \textbf{17.22} & \textbf{0.280} 
            & \textbf{12.17} & \textbf{0.216} 
            & \textbf{7.56} & \textbf{0.139} 
            & \textbf{6.33} & \textbf{0.106} 
            & \textbf{5.73} & \textbf{0.082} 
            & \textbf{5.93} & \textbf{0.078} 
            & \textbf{8.03} & \textbf{0.029} \\

            BAMCTS
            & 43.47 & 0.488 & 38.65 & 0.433 & 28.92 & 0.318 & 22.24 & 0.244 & 16.28 & 0.181 & 13.69 & 0.157 & 24.38 & 0.063 \\

            \bottomrule
        \end{tabular}%
    }
\end{table}

For density tracking, COMBO achieves the lowest average RMSE of 0.691, with MOPO (0.727) and PPO (0.733) close behind. 
The relatively small gap between model-based and model-free methods suggests that the offline dataset provides sufficient coverage for learning effective density-control behavior. 
By contrast, RAMBO performs noticeably worse, with an average RMSE of 1.583, indicating that robustness-oriented model-based training does not consistently improve performance across all profile control tasks.

\begin{table}[htbp]
    \vspace{-0.5em}
    \centering
    \caption{Density profile tracking error (RMSE $\downarrow$). Values are reported as mean $\pm$ standard error over test shots.}
    \label{tab:density}

    \resizebox{\textwidth}{!}{%
        \begin{tabular}{lcccccccccccccc}
            \toprule
            \multirow{2}{*}{\textbf{Algorithm}} & \multicolumn{2}{c}{$\psi_{0.09}$} & \multicolumn{2}{c}{$\psi_{0.18}$} & \multicolumn{2}{c}{$\psi_{0.39}$} & \multicolumn{2}{c}{$\psi_{0.58}$} & \multicolumn{2}{c}{$\psi_{0.79}$} & \multicolumn{2}{c}{$\psi_{0.88}$} & \multicolumn{2}{c}{Average} \\
            \cmidrule(lr){2-3} \cmidrule(lr){4-5} \cmidrule(lr){6-7} \cmidrule(lr){8-9} \cmidrule(lr){10-11} \cmidrule(lr){12-13} \cmidrule(lr){14-15}
             & Mean & SE & Mean & SE & Mean & SE & Mean & SE & Mean & SE & Mean & SE & Mean & SE \\
            \midrule

            GCIL
            & 1.120 & 0.0135 & 1.031 & 0.0128 & 0.920 & 0.0121 & 0.864 & 0.0117 & 0.792 & 0.0115 & 0.760 & 0.0113 & 0.860 & 0.0021 \\

            \midrule

            TD3BC & 1.093 & 0.0132 & 1.009 & 0.0126 & 0.900 & 0.0117 & 0.840 & 0.0113 & 0.761 & 0.0109 & 0.727 & 0.0107 & 0.836 & 0.0021 \\
           
            CQL & 1.127 & 0.0140 & 1.033 & 0.0132 & 0.922 & 0.0123 & 0.868 & 0.0120 & 0.807 & 0.0120 & 0.775 & 0.012 & 0.865 & 0.0022 \\

            IQL
            & 1.088 & 0.0130 & 0.998 & 0.0124 & 0.885 & 0.0118 & 0.830 & 0.0117 & 0.768 & 0.0117 & 0.738 & 0.0116 & 0.831 & 0.0021 \\
           
            EDAC & 1.281 & 0.0157 & 1.197 & 0.0151 & 1.099 & 0.0144 & 1.051 & 0.0142 & 0.998 & 0.0145 & 0.969 & 0.0145 & 1.037 & 0.0026 \\

            MCQ
            & 1.059 & 0.0119 & 0.983 & 0.0115 & 0.891 & 0.0109 & 0.840 & 0.0106 & 0.768 & 0.0104 & 0.735 & 0.0102 & 0.828 & 0.0019 \\

            \midrule
            
            PPO & 1.048 & 0.0131 & 0.914 & 0.0125 & 0.770 & 0.0115 & 0.717 & 0.0108 & 0.661 & 0.0100 & 0.630 & 0.0096 & 0.733 & 0.0020 \\
           
            COMBO 
            & \textbf{0.966} & \textbf{0.0115} 
            & \textbf{0.870} & \textbf{0.0107} 
            & \textbf{0.745} & \textbf{0.0098} 
            & \textbf{0.676} & \textbf{0.0094} 
            & \textbf{0.602} & \textbf{0.0091} 
            & \textbf{0.578} & \textbf{0.0088} 
            & \textbf{0.691} & \textbf{0.0017} \\
           
            MOPO & 0.988 & 0.0108 & 0.898 & 0.0105 & 0.784 & 0.0103 & 0.723 & 0.0102 & 0.648 & 0.0096 & 0.621 & 0.0093 & 0.727 & 0.0018 \\
           
            MOBILE & 1.325 & 0.0144 & 1.273 & 0.0139 & 1.225 & 0.0135 & 1.206 & 0.0137 & 1.195 & 0.0143 & 1.197 & 0.0145 & 1.184 & 0.0025 \\

            RAMBO
            & 1.576 & 0.0147 & 1.596 & 0.0138 & 1.671 & 0.0132 & 1.699 & 0.0127 & 1.638 & 0.0115 & 1.600 & 0.0109 & 1.583 & 0.0024 \\

            BAMCTS
            & 1.220 & 0.0143 & 1.165 & 0.0142 & 1.101 & 0.0139 & 1.058 & 0.0135 & 0.975 & 0.0128 & 0.938 & 0.0125 & 1.023 & 0.0024 \\

            \bottomrule
        \end{tabular}%
    }
\end{table}

For temperature tracking, MOPO achieves the best average RMSE (0.193), outperforming PPO (0.240), MCQ (0.244), and COMBO (0.264). Temperature dynamics involve delayed responses and accumulated transport effects, making long-horizon accuracy particularly important. Methods that penalize uncertain model predictions (MOPO) are well suited to this regime, as they avoid compounding errors from unreliable model regions.

\begin{table}[htbp]
    \centering
    \caption{Temperature profile tracking error (RMSE $\downarrow$). Values are reported as mean $\pm$ standard error over test shots.}
    \label{tab:temperature}

    \resizebox{\textwidth}{!}{%
        \begin{tabular}{lcccccccccccccc}
            \toprule
            \multirow{2}{*}{\textbf{Algorithm}} & \multicolumn{2}{c}{$\psi_{0.09}$} & \multicolumn{2}{c}{$\psi_{0.18}$} & \multicolumn{2}{c}{$\psi_{0.39}$} & \multicolumn{2}{c}{$\psi_{0.58}$} & \multicolumn{2}{c}{$\psi_{0.79}$} & \multicolumn{2}{c}{$\psi_{0.88}$} & \multicolumn{2}{c}{Average} \\
            
            \cmidrule(lr){2-3} \cmidrule(lr){4-5} \cmidrule(lr){6-7}
            \cmidrule(lr){8-9} \cmidrule(lr){10-11} \cmidrule(lr){12-13}
            \cmidrule(lr){14-15}

             & Mean & SE & Mean & SE & Mean & SE
            & Mean & SE & Mean & SE & Mean & SE
            & Mean & SE \\
            \midrule

            GCIL
            & 0.492 & 0.0055 & 0.461 & 0.0052 & 0.393 & 0.0045 & 0.325 & 0.0037 & 0.238 & 0.0028 & 0.206 & 0.0025 & 0.323 & 0.0008 \\

            \midrule

            TD3BC
            & 0.468 & 0.0051 & 0.429 & 0.0047 & 0.360 & 0.0039 & 0.298 & 0.0033 & 0.223 & 0.0026 & 0.195 & 0.0024 & 0.301 & 0.0007 \\

            CQL
            & 0.563 & 0.0054 & 0.526 & 0.0049 & 0.443 & 0.0041 & 0.360 & 0.0034 & 0.259 & 0.0026 & 0.223 & 0.0023 & 0.362 & 0.0008 \\

            IQL
            & 0.465 & 0.0052 & 0.429 & 0.0049 & 0.362 & 0.0041 & 0.301 & 0.0034 & 0.225 & 0.0027 & 0.196 & 0.0025 & 0.302 & 0.0007 \\

            EDAC
            & 0.735 & 0.0073 & 0.709 & 0.0070 & 0.624 & 0.0061 & 0.518 & 0.0053 & 0.376 & 0.0043 & 0.330 & 0.0041 & 0.507 & 0.0011 \\

            MCQ
            & 0.382 & 0.0046 & 0.349 & 0.0043 & 0.293 & 0.0037 & 0.241 & 0.0030 & 0.178 & 0.0023 & 0.157 & 0.0021 & 0.244 & 0.0006 \\

            \midrule
            
            PPO
            & 0.444 & 0.0040 & 0.362 & 0.0034 & 0.245 & 0.0029 & 0.204 & 0.0026 & 0.187 & 0.0025 & 0.179 & 0.0026 & 0.240 & 0.0006 \\

            COMBO
            & 0.408 & 0.0038 & 0.371 & 0.0035 & 0.310 & 0.0030 & 0.261 & 0.0026 & 0.203 & 0.0022 & 0.180 & 0.0021 & 0.264 & 0.0006 \\

            MOPO
            & \textbf{0.276} & \textbf{0.0032} 
            & \textbf{0.246} & \textbf{0.0031} 
            & \textbf{0.212} & \textbf{0.0030} 
            & \textbf{0.190} & \textbf{0.0030} 
            & \textbf{0.163} & \textbf{0.0030} 
            & \textbf{0.156} & \textbf{0.0030} 
            & \textbf{0.193} & \textbf{0.0005} \\

            MOBILE
            & 0.623 & 0.0066 & 0.580 & 0.0062 & 0.488 & 0.0054 & 0.403 & 0.0044 & 0.304 & 0.0033 & 0.273 & 0.0030 & 0.408 & 0.0009 \\

            RAMBO
            & 1.044 & 0.0105 & 1.004 & 0.0100 & 0.848 & 0.0087 & 0.652 & 0.0071 & 0.406 & 0.0048 & 0.332 & 0.0041 & 0.655 & 0.0016 \\

            BAMCTS
            & 0.814 & 0.0115 & 0.770 & 0.0107 & 0.658 & 0.0089 & 0.537 & 0.0072 & 0.383 & 0.0052 & 0.330 & 0.0046 & 0.534 & 0.0015 \\

            \bottomrule
        \end{tabular}%
    }
\end{table}

Pressure tracking is likely the most strongly coupled task. The pressure profile depends directly on both density and temperature, and it is also influenced by the current profile and actuator dynamics. In this task, MOPO achieves the lowest average RMSE of 5198.6, followed by RAMBO at 5358.2. Among model-free methods, EDAC performs best with an average RMSE of 6016.8. These results suggest that uncertainty-aware methods may be better suited to pressure tracking. 

\begin{table}[htbp]
\vspace{-0.5em}
    \centering
    \caption{Pressure profile tracking error (RMSE $\downarrow$). Values are reported as mean $\pm$ standard error over test shots.}
    \label{tab:pressure}

    \resizebox{\textwidth}{!}{%
        \begin{tabular}{lcccccccccccccc}
            \toprule
            \multirow{2}{*}{\textbf{Algorithm}} & \multicolumn{2}{c}{$\psi_{0.09}$} & \multicolumn{2}{c}{$\psi_{0.18}$} & \multicolumn{2}{c}{$\psi_{0.39}$} & \multicolumn{2}{c}{$\psi_{0.58}$} & \multicolumn{2}{c}{$\psi_{0.79}$} & \multicolumn{2}{c}{$\psi_{0.88}$} & \multicolumn{2}{c}{Average} \\
            
            \cmidrule(lr){2-3} \cmidrule(lr){4-5} \cmidrule(lr){6-7}
            \cmidrule(lr){8-9} \cmidrule(lr){10-11} \cmidrule(lr){12-13}
            \cmidrule(lr){14-15}

             & Mean & SE & Mean & SE & Mean & SE
            & Mean & SE & Mean & SE & Mean & SE
            & Mean & SE \\
            \midrule

            GCIL & 17346.5 & 177.2 & 15052.4 & 160.0 & 10897.1 & 129.6 & 8174.0 & 103.5 & 4943.7 & 64.2 & 3300.2 & 42.8 & 8775.3 & 25.4 \\

            \midrule

            TD3BC
            & 15496.3 & 154.0 & 13049.8 & 133.2 & 9079.8 & 108.5 & 6998.2 & 92.6 & 4558.2 & 61.6 & 3138.1 & 42.1 & 7628.9 & 21.8 \\
            
            CQL
            & 13685.7 & 144.8 & 11030.1 & 123.6 & 6921.7 & 92.8 & 5275.7 & 74.2 & 3614.5 & 48.4 & 2544.5 & 33.1 & 6162.2 & 19.1 \\

            IQL & 15518.4 & 171.7 & 13089.6 & 153.7 & 9041.5 & 126.1 & 6788.6 & 104.7 & 4312.7 & 66.9 & 2951.4 & 45.0 & 7525.8 & 23.8 \\
            
            EDAC
            & 12405.6 & 139.3 & 9868.5 & 114.9 & 6624.9 & 81.4 & 5648.2 & 68.9 & 4114.2 & 48.9 & 2922.3 & 34.4 & 6016.8 & 17.2 \\

            MCQ & 14885.6 & 149.0 & 12482.0 & 129.0 & 8536.6 & 101.2 & 6457.9 & 83.5 & 4155.5 & 54.1 & 2852.5 & 36.8 & 7174.3 & 20.8 \\

            \midrule
            
            PPO
            & 14511.3 & 143.5 & 11680.0 & 117.8 & 7149.0 & 80.1 & 5178.8 & 63.5 & 3511.9 & 41.1 & 2482.2 & 28.2 & 6333.2 & 18.8 \\
            
            COMBO
            & 14014.7 & 135.6 & 11379.7 & 110.5 & 7002.4 & 73.7 & 4959.2 & 56.6 & 3215.5 & 36.8 & 2242.7 & 25.5 & 6095.9 & 18.0 \\
            
            MOPO
            & \textbf{12706.3} & \textbf{131.5} 
            & \textbf{9894.5} & \textbf{105.4} 
            & \textbf{5534.0} & \textbf{66.7} 
            & \textbf{3997.4} & \textbf{52.9} 
            & \textbf{2892.6} & \textbf{37.4} 
            & \textbf{2094.5} & \textbf{26.5} 
            & \textbf{5198.6} & \textbf{16.5} \\
            
            MOBILE
            & 15342.5 & 150.7 & 12692.0 & 125.8 & 8178.6 & 90.7 & 5738.4 & 75.3 & 3464.5 & 51.9 & 2353.1 & 36.1 & 6854.9 & 20.7 \\

            RAMBO & 12437.8 & 133.2 & 9786.2 & 109.9 & 5790.4 & 77.2 & 4391.8 & 63.4 & 3171.1 & 43.2 & 2276.1 & 30.0 & 5358.2 & 17.0 \\

            BAMCTS & 18697.5 & 180.6 & 16093.4 & 152.6 & 11559.1 & 119.0 & 8715.9 & 103.0 & 5420.3 & 69.6 & 3673.5 & 47.7 & 9411.5 & 25.6 \\

            \bottomrule
        \end{tabular}%
    }
\end{table}

Methods without such stabilization mechanisms tend to obtain higher errors.

Across all the tasks, MOPO is the most robust method; COMBO is strongest on density; RAMBO excels on rotation and pressure but fails on density and temperature; PPO is a strong baseline on rotation, density, and temperature, but falls short on the more coupled pressure profile tracking task. Thus, no single algorithm is universally beneficial across plasma profile control objectives, calling for stronger offline RL algorithms.

\begin{figure}[htpb]
\vspace{-1.0em}
  \centering
  \begin{subfigure}{\textwidth}
    \centering
    \includegraphics[width=\linewidth]{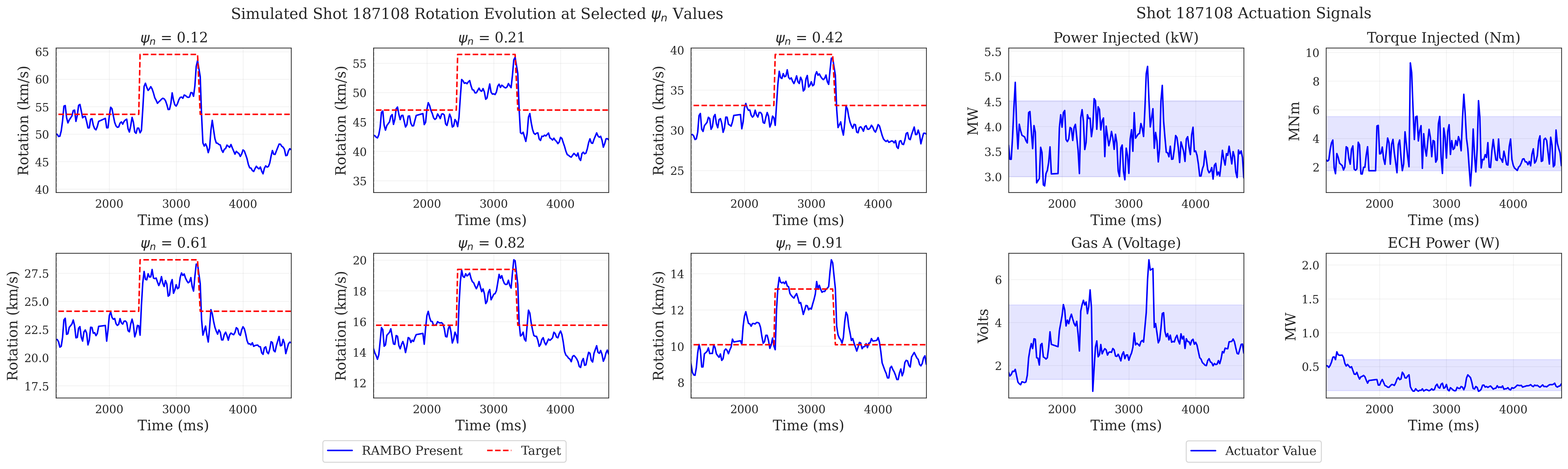}
    \caption{Higher rotation target}
  \end{subfigure}
   \vspace{-0.4em}
  \begin{subfigure}{\textwidth}
    \centering
    \includegraphics[width=\linewidth]{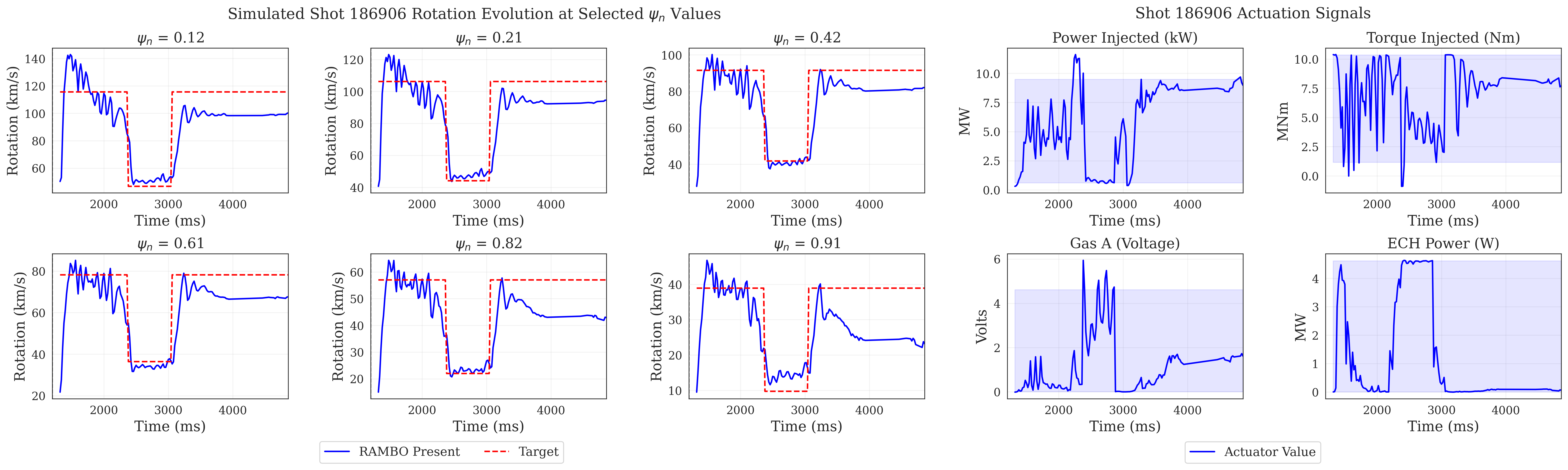}
    \caption{Lower rotation target}
  \end{subfigure}

  \caption{Simulated results of RAMBO applied to Rotation profile tracking for shot 187108 and 186906 using the dynamics-model environment. Two target patterns are tested: (a) increasing the profile and returning, and (b) decreasing the profile and returning. Left plots show the rotation profile at different normalized flux values ($\psi_{\mathrm{n}}$). Right plots show the RL-controlled actuator signals. Both cases demonstrate strong tracking performance in the absence of the sim-to-real gap.}
  \label{fig:rot}
\end{figure}

A second consistent trend is that tracking near the plasma core remains more difficult than tracking near the edge. Errors at smaller normalized flux values are systematically higher regardless of the algorithm or task, because both the profile magnitude and temporal variation are substantially larger near the core. This tendency is also qualitatively reflected in Appendix~\ref{app:additionalresults}.

\begin{figure}[htpb]
  \vspace{-1.0em}
  \centering

  \begin{subfigure}{\textwidth}
    \centering
    \includegraphics[width=\linewidth]{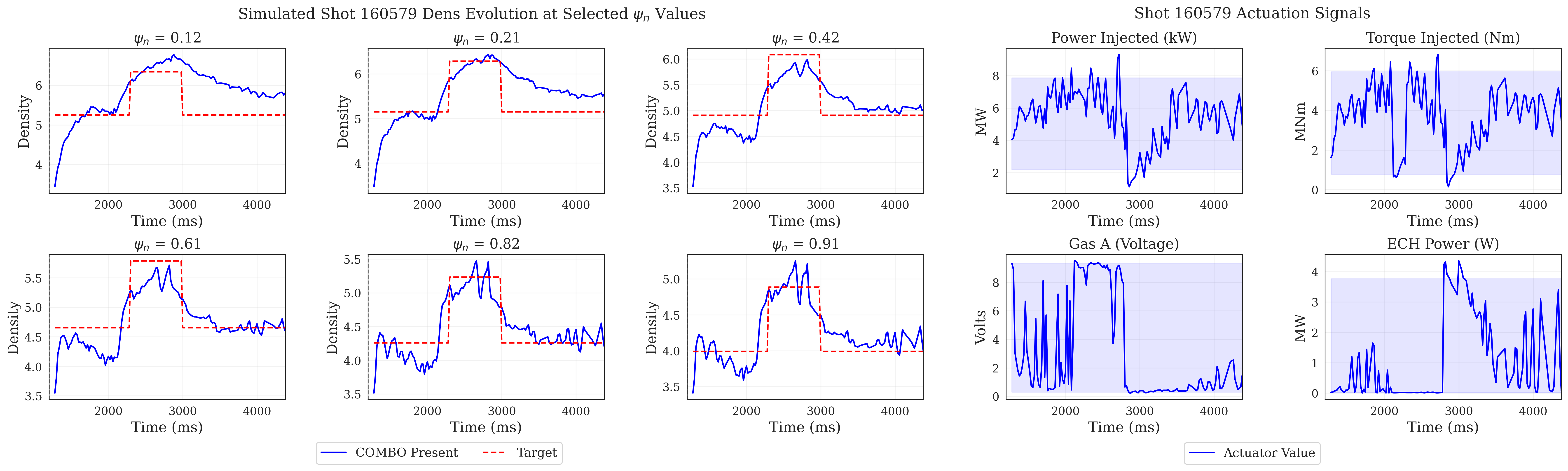}
    \caption{Higher density target}
  \end{subfigure}
 
  \begin{subfigure}{\textwidth}
    \centering
    \includegraphics[width=\linewidth]{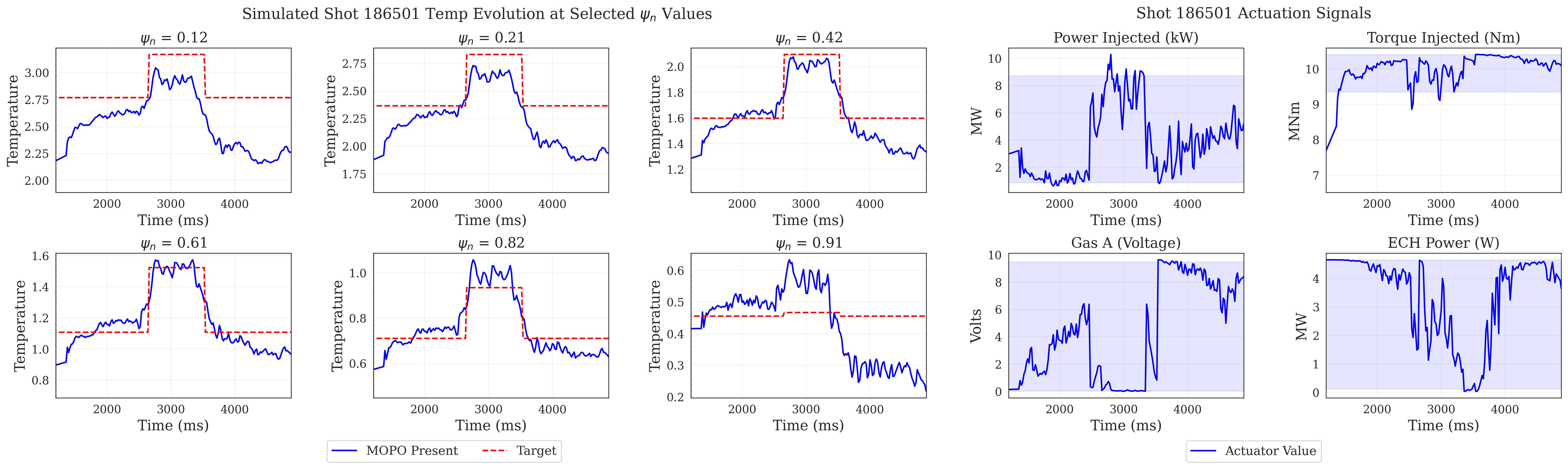}
    \caption{Higher temperature target}
  \end{subfigure}
  
  \begin{subfigure}{\textwidth}
    \centering
    \includegraphics[width=\linewidth]{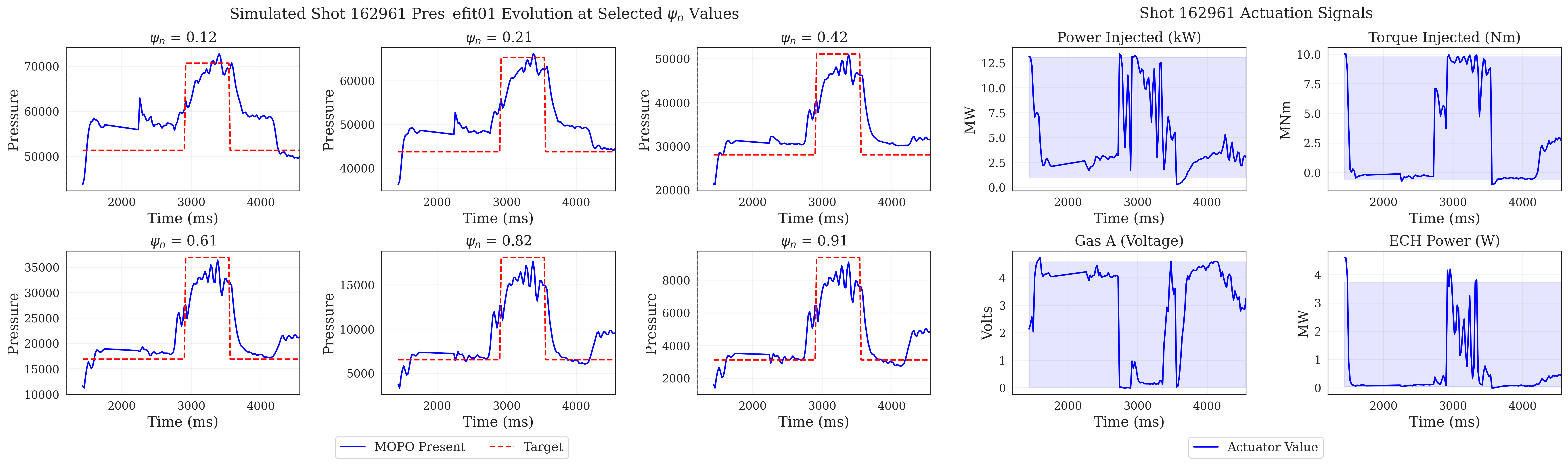}
    
    \caption{Higher pressure target}
  \end{subfigure}

  \caption{Simulated results for the higher-target cases of the remaining three tasks in the dynamics-model environment. From top to bottom, the panels show COMBO on density profile tracking (shot 160579), MOPO on temperature profile tracking (shot 186501), and MOPO on pressure profile tracking (shot 162961).}
  \label{fig:other_tasks}
\end{figure}

The small standard errors across all tables indicate that performance is consistent across test shots and that the observed gaps reflect systematic differences between methods rather than noise or a few favorable trajectories.

The main text shows higher- and lower-target rotation rollouts (Fig.~\ref{fig:rot}), as well as higher-target rollouts for density, temperature, and pressure (Fig.~\ref{fig:other_tasks}). The corresponding lower-target cases for density, temperature, and pressure are provided in Appendix~\ref{app:additionalresults}. Across all tasks, the policies exhibit a consistent closed-loop pattern: each target switch triggers an immediate coordinated response across all four actuators, after which the profile converges smoothly to the new setpoint. The dominant actuation pathway differs by task: rotation is driven primarily by torque, density by gas puffing, temperature by co-dominant ECH and beam power with active gas puffing suppression, and pressure by balanced coordination of all four actuators. Overall, although there is a consistent offset in the tracking values, the controller responded appropriately to each step change, coordinating the actuators in a physically meaningful way.

\section{Conclusion and Discussion}

\label{sec:discuss}
We introduced RL4F, an open-source benchmark for offline RL in nuclear-fusion plasma profile control, releasing reference dynamics models trained on real-world fusion operational data, synthetic datasets for offline RL training, and the full codebase to support reproducibility and community adoption. The benchmark provides standardized task definitions and a unified evaluation protocol across four profile-tracking objectives: rotation, density, temperature, and pressure. Physics-based simulators such as RAPTOR~\citep{felici2011realtime_raptor} and TORAX~\citep{torax2024arxiv} can struggle to capture complex fusion dynamics, and calibrating them to a specific tokamak can be nontrivial. In contrast, our benchmark is built directly from experimental discharge data and designed specifically for offline RL evaluation. To the best of our knowledge, it is the first offline RL benchmark for nuclear-fusion plasma control, providing a common testbed for comparing algorithms under fixed datasets and standardized evaluation conditions.

A current limitation is that RL4F is constructed from DIII-D data alone. As a result, it remains unclear how well the learned dynamics model, or the relative performance of different algorithms, transfers to other tokamak devices. Addressing this limitation will require broader collaboration across the fusion community and the inclusion of data from additional fusion devices. Despite these limitations, we believe this benchmark is a useful step toward accelerating progress in both nuclear fusion and RL algorithm development.


\bibliographystyle{apalike}
\bibliography{ref}

\newpage
\appendix
\newcommand{\cmark}{\textcolor{green!60!black}{\ding{51}}}
\newcommand{\xmark}{\textcolor{red}{\ding{55}}}

\section{Additional Results}
\label{app:additionalresults}
\begin{figure}[htpb]
  \centering

  \begin{subfigure}{1\textwidth}
    \centering
    \includegraphics[width=\linewidth]{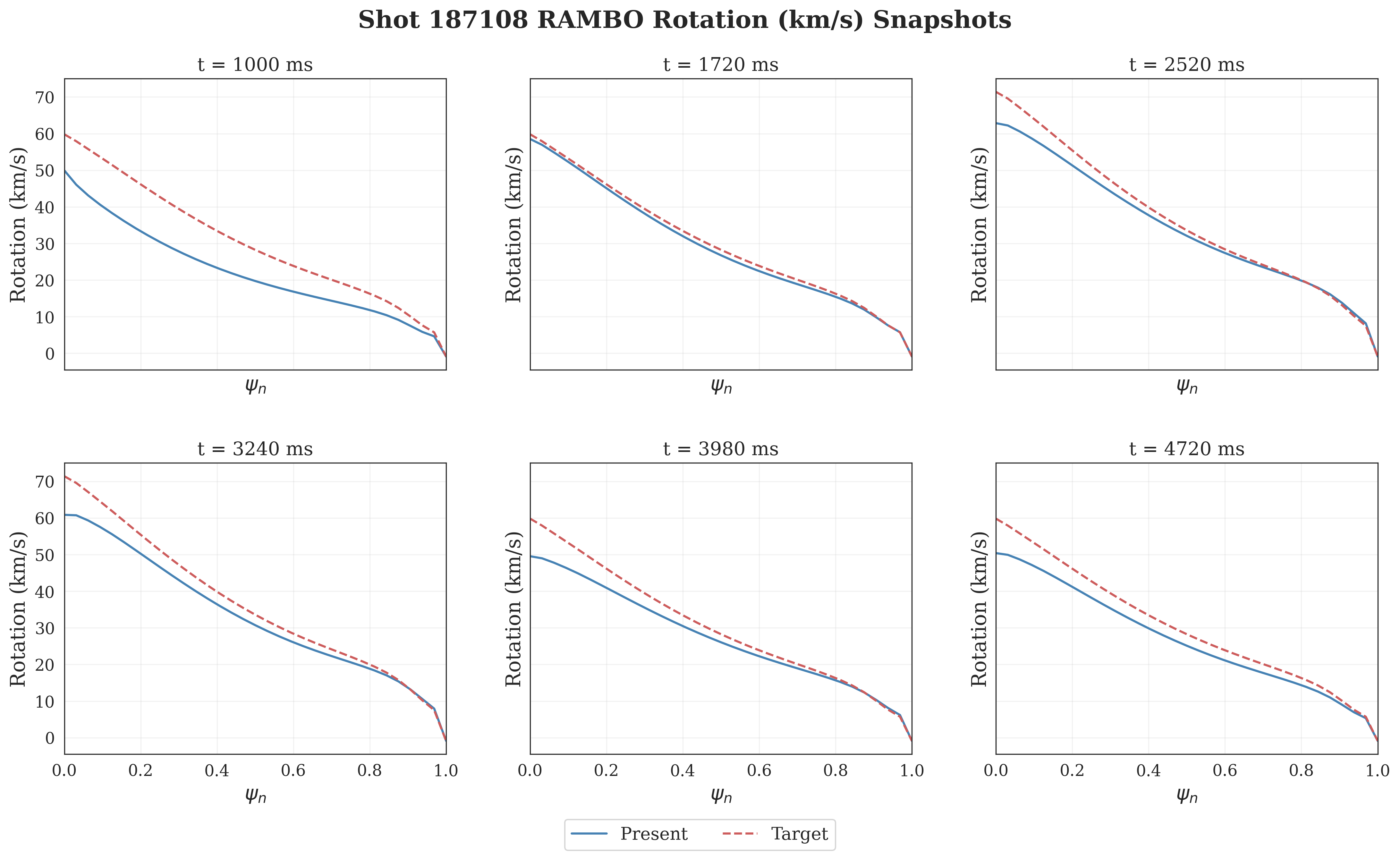}
    \caption{Higher rotation target}
  \end{subfigure}

  \vspace{0.5em}

  \begin{subfigure}{1\textwidth}
    \centering
    \includegraphics[width=\linewidth]{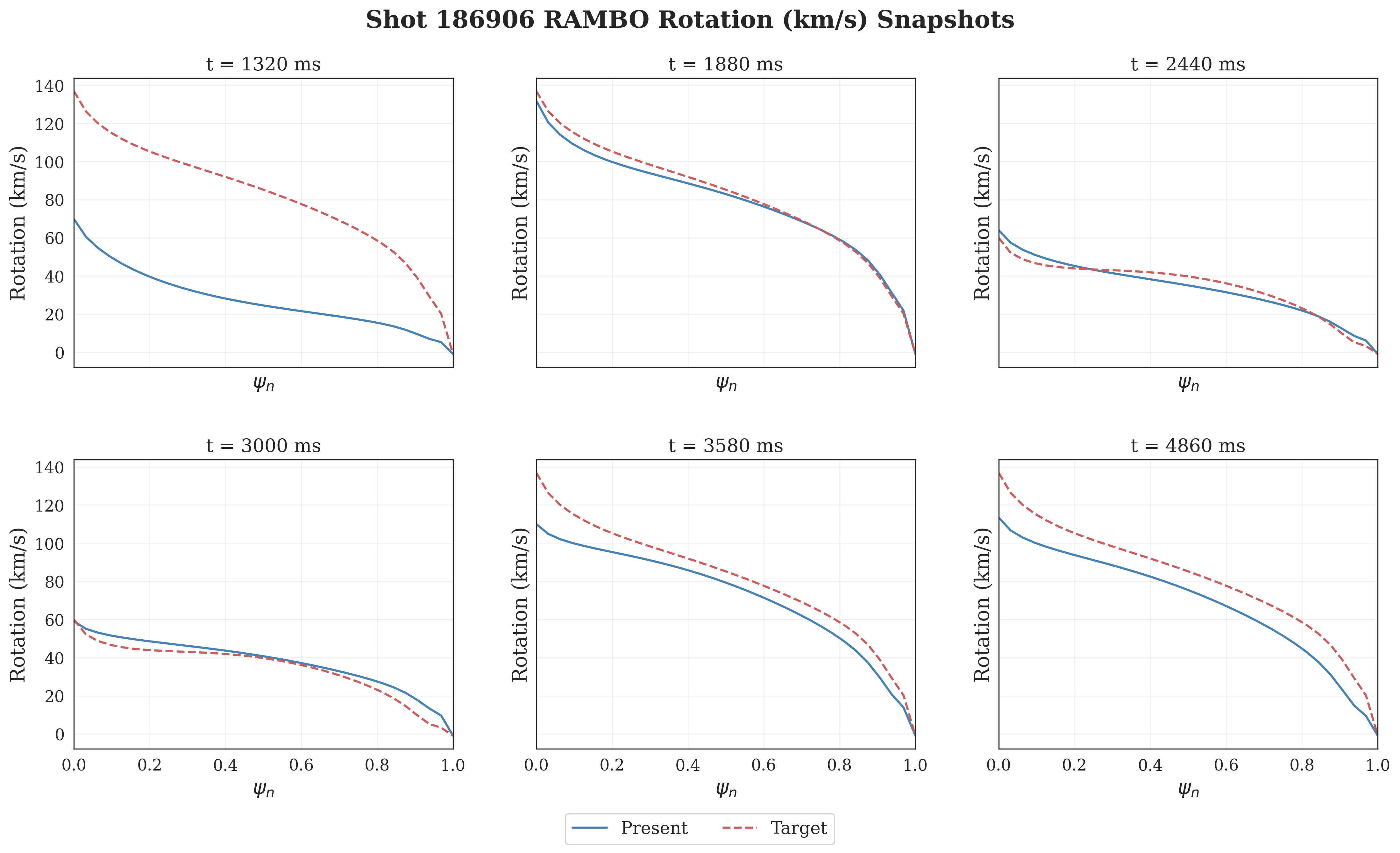}
    \caption{Lower rotation target}
  \end{subfigure}

  \caption{Temporal evolution of full rotation profiles at selected time instances for two representative shots using RAMBO. (a) Higher rotation target and (b) Lower rotation target. Solid lines represent the present (predicted) profiles, while dashed lines indicate the target profiles.}
  \label{fig:rot_full}
\end{figure}

\begin{figure}[htpb]
  \centering

  \begin{subfigure}{1\textwidth}
    \centering
    \includegraphics[width=\linewidth]{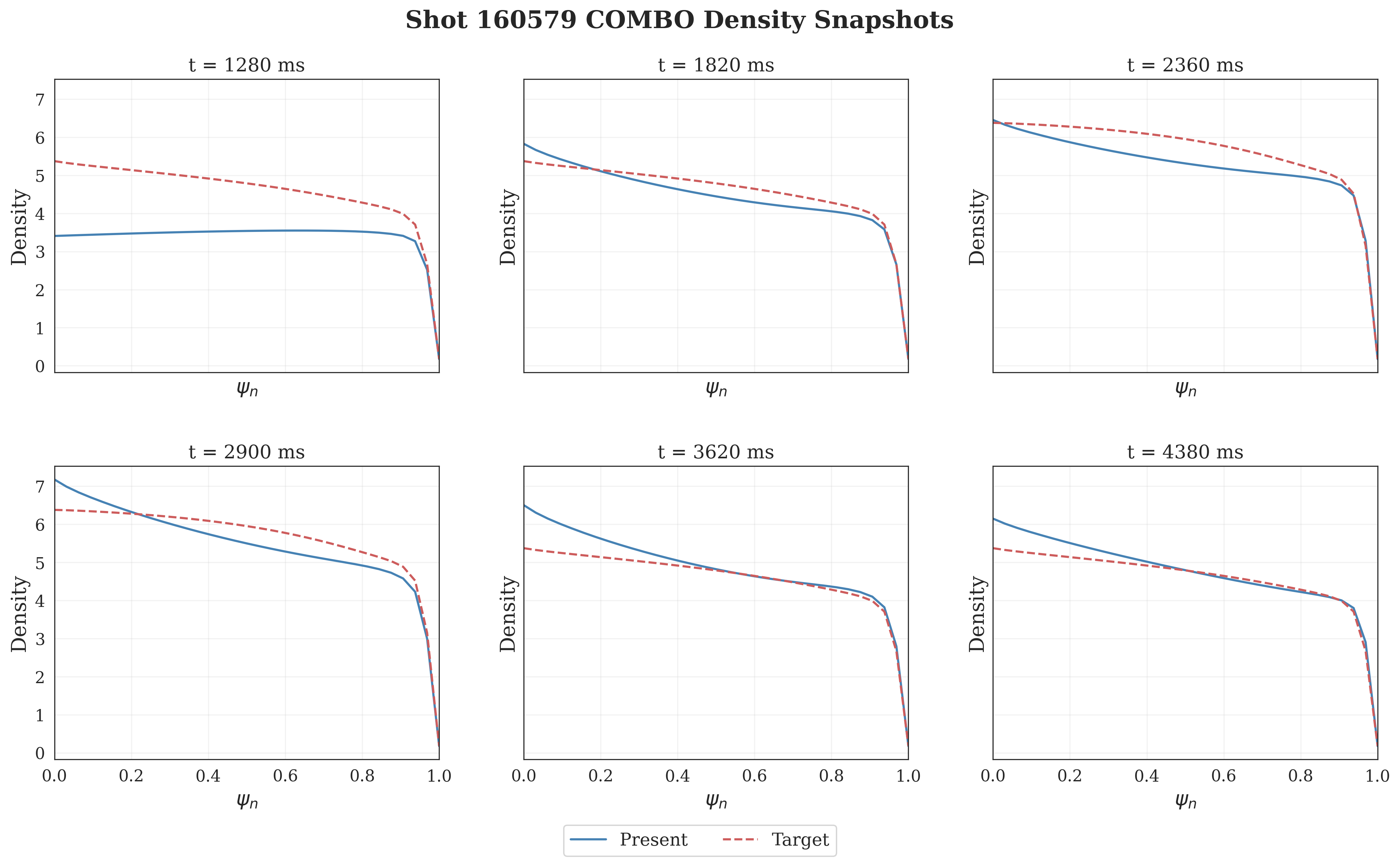}
    \caption{Higher density target}
  \end{subfigure}

  \vspace{0.5em}

  \begin{subfigure}{1\textwidth}
    \centering
    \includegraphics[width=\linewidth]{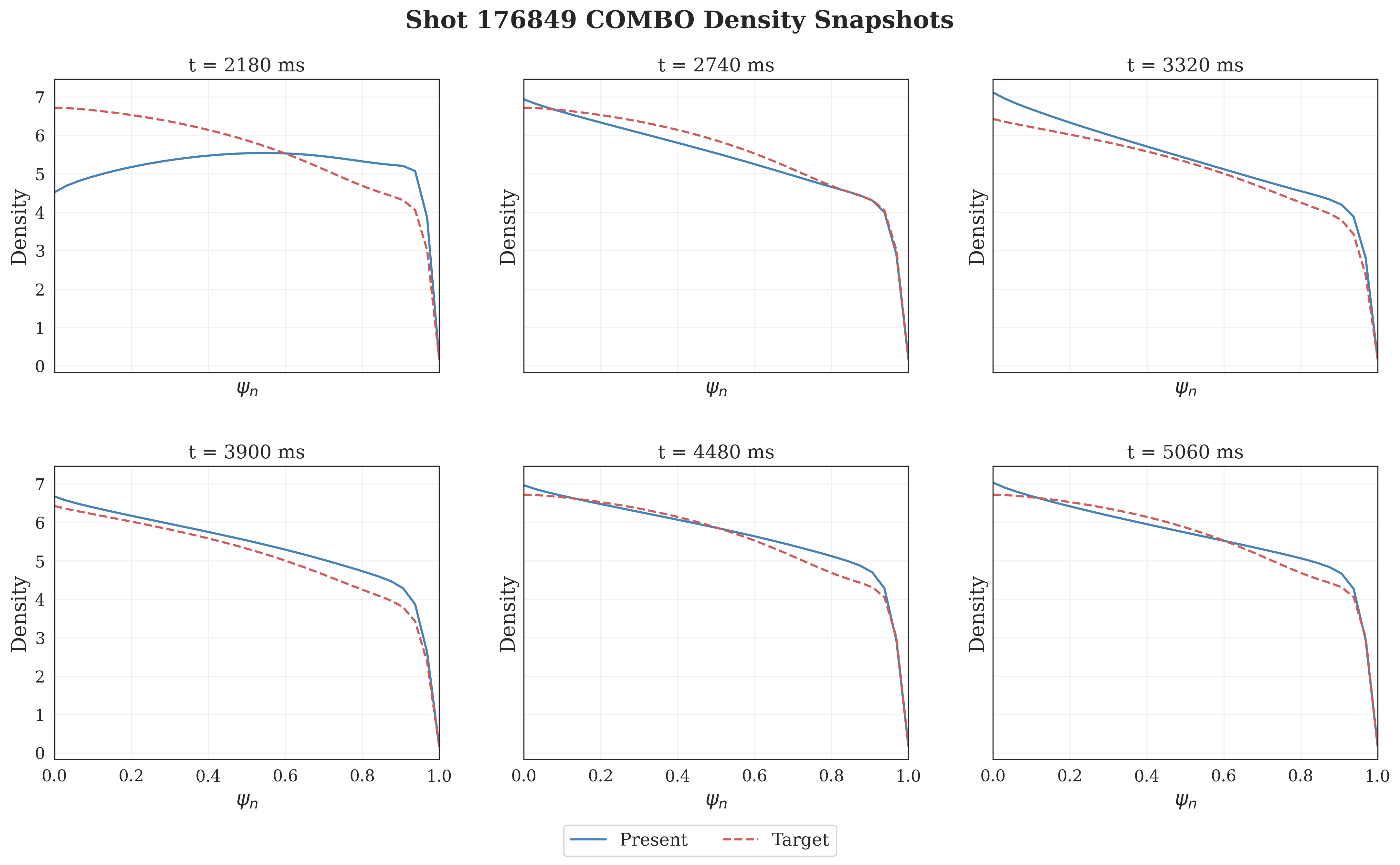}
    \caption{Lower density target}
  \end{subfigure}

  \caption{Temporal evolution of full density profiles at selected time instances for two representative shots using COMBO. (a) Higher density target and (b) Lower density target. Solid lines represent the present (predicted) profiles, while dashed lines indicate the target profiles.}
  \label{fig:dens_full}
\end{figure}

\begin{figure}
  \centering

  \begin{subfigure}{1\textwidth}
    \centering
    \includegraphics[width=\linewidth]{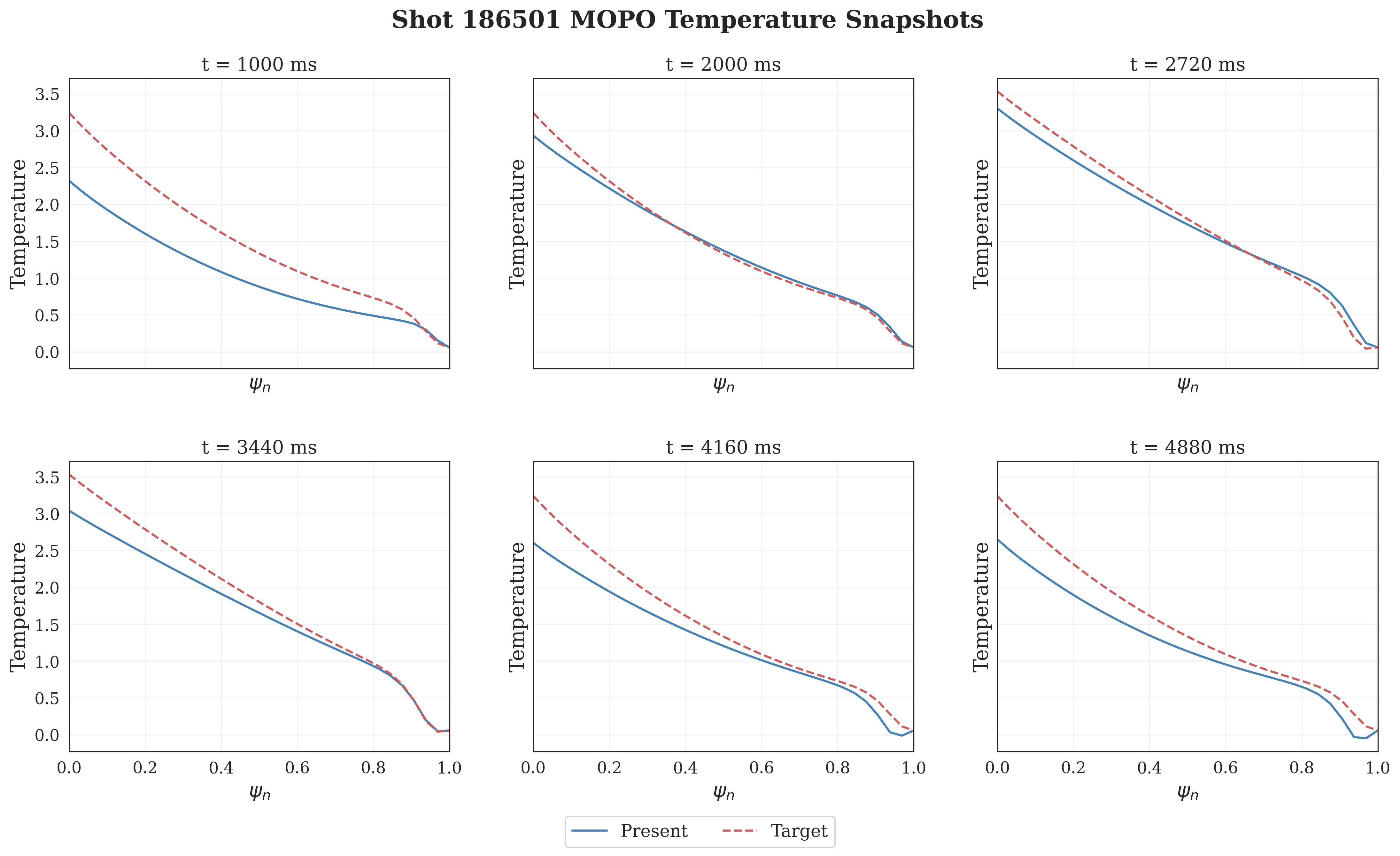}
    \caption{Higher temperature target}
  \end{subfigure}

  \vspace{0.5em}

  \begin{subfigure}{1\textwidth}
    \centering
    \includegraphics[width=\linewidth]{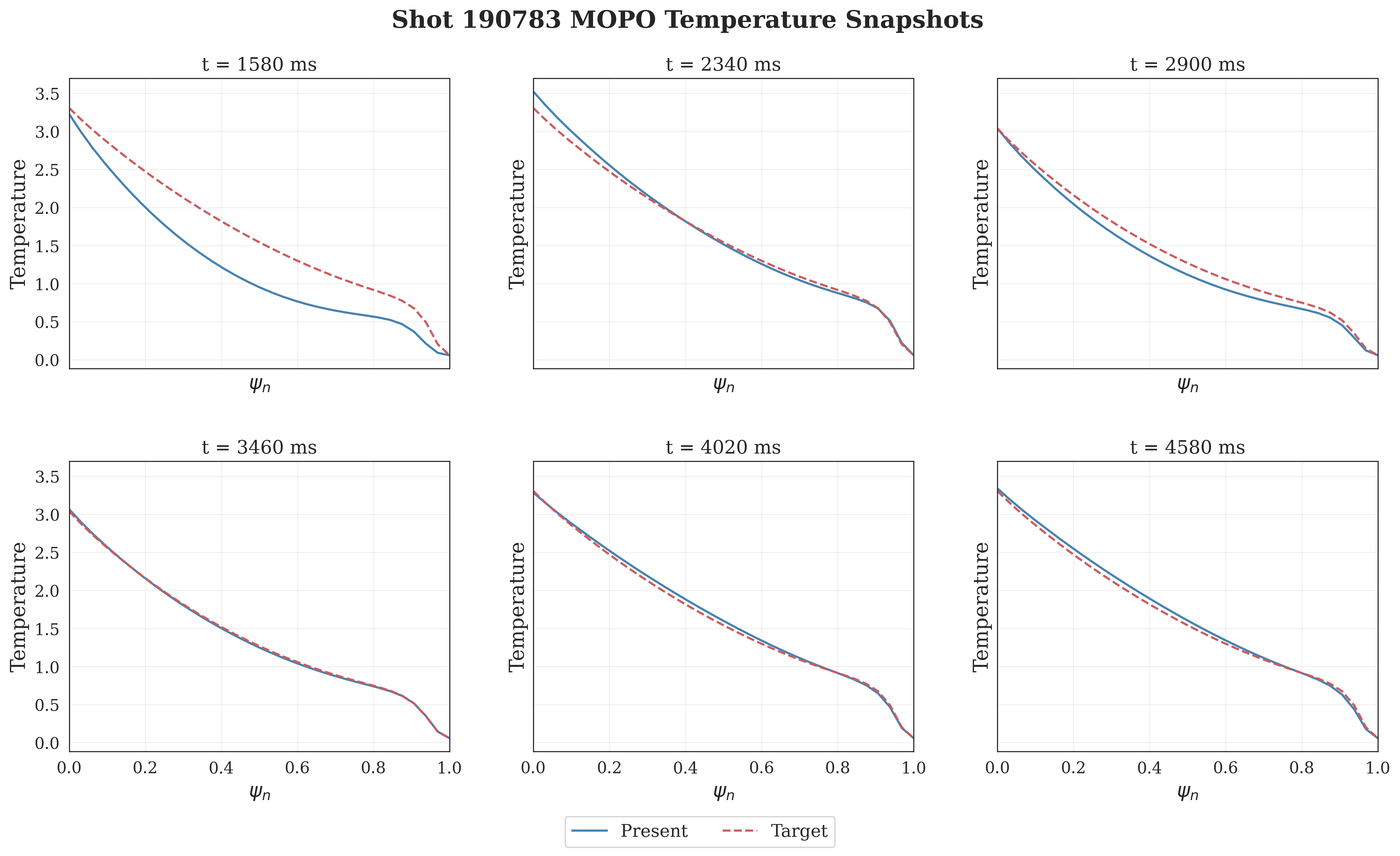}
    \caption{Lower temperature target}
  \end{subfigure}

  \caption{Temporal evolution of full temperature profiles at selected time instances for two representative shots using MOPO. (a) Higher temperature target and (b) Lower temperature target. Solid lines represent the present (predicted) profiles, while dashed lines indicate the target profiles.}
  \label{fig:temp_full}
\end{figure}

\begin{figure}
  \centering

  \begin{subfigure}{1\textwidth}
    \centering
    \includegraphics[width=\linewidth]{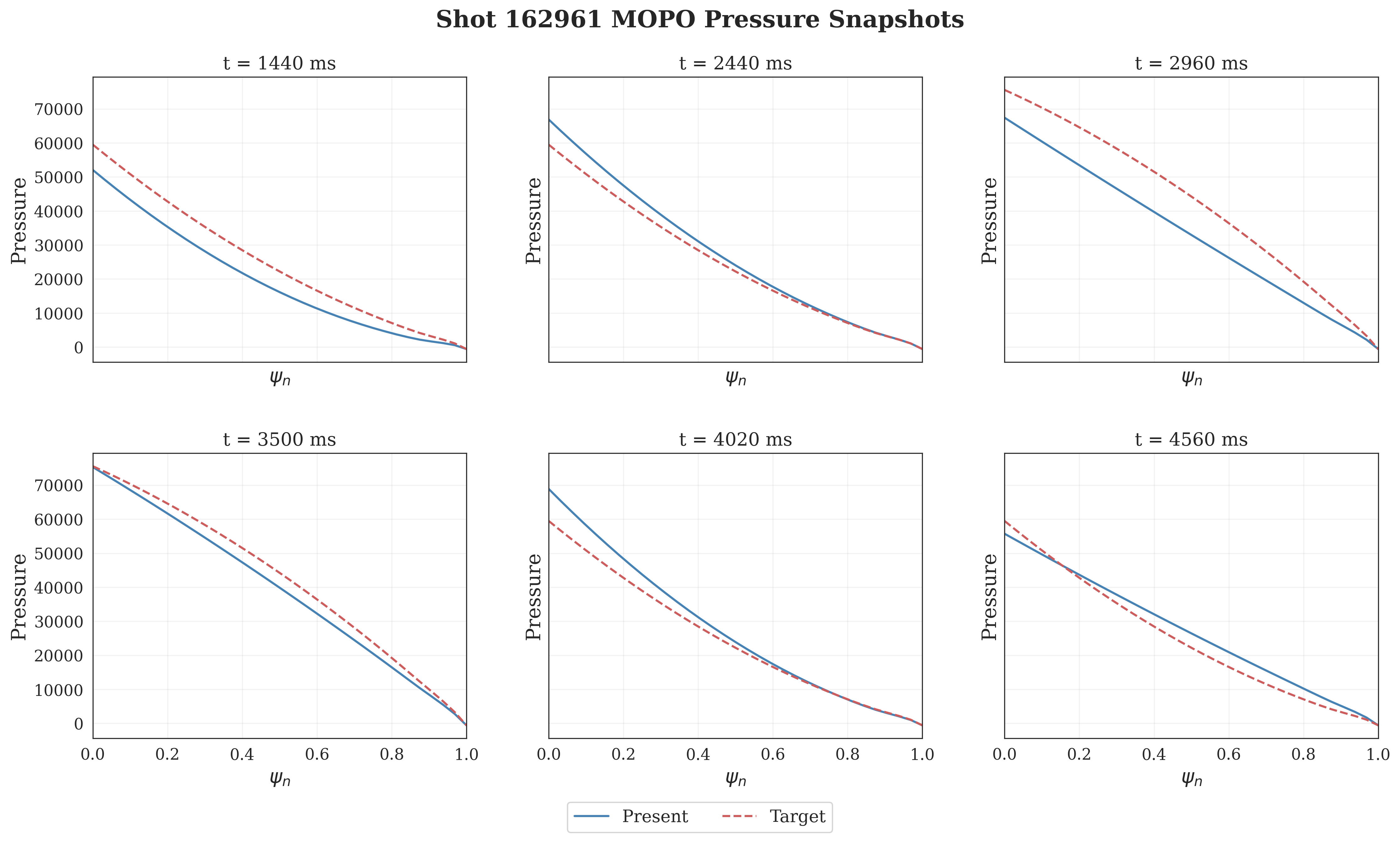}
    \caption{Higher pressure target}
  \end{subfigure}

  \vspace{0.5em}

  \begin{subfigure}{1\textwidth}
    \centering
    \includegraphics[width=\linewidth]{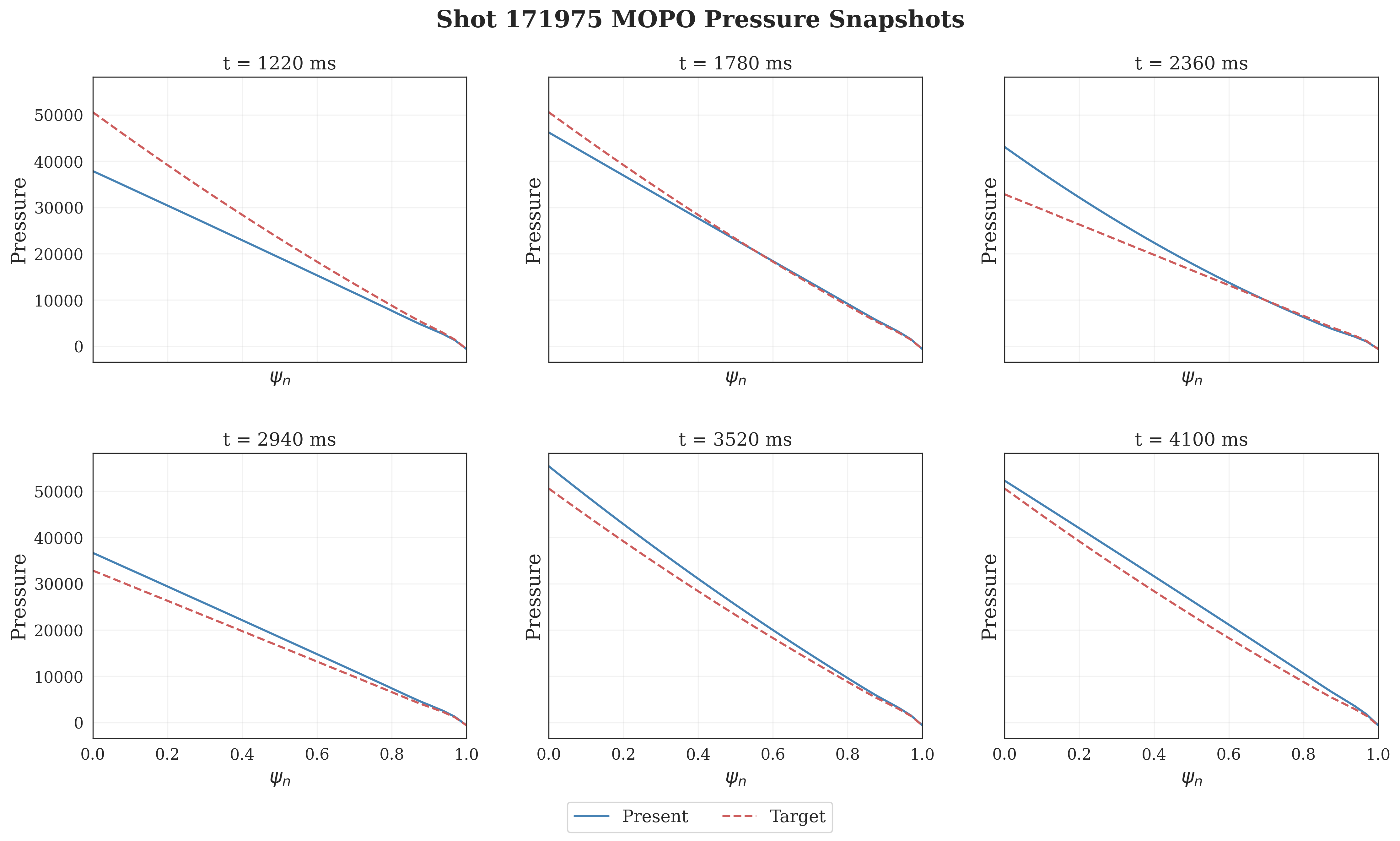}
    \caption{Lower pressure target}
  \end{subfigure}

  \caption{Temporal evolution of full pressure profiles at selected time instances for two representative shots using MOPO. (a) Higher pressure target and (b) Lower pressure target. Solid lines represent the present (predicted) profiles, while dashed lines indicate the target profiles.}
  \label{fig:pres_full}
\end{figure}

We provide additional visualizations of closed-loop profile tracking trajectories for each of the four benchmark tasks. Figures~\ref{fig:rot_full}-~\ref{fig:pres_full} show the temporal evolution of the full radial profiles at selected time instances for two representative test shots per task, corresponding to a higher and a lower target profile, respectively. In each figure, solid lines denote the model-predicted profile and dashed lines indicate the target profile.

\begin{figure}[htpb]
  \centering

  \begin{subfigure}{\textwidth}
    \centering
    \includegraphics[width=\linewidth]{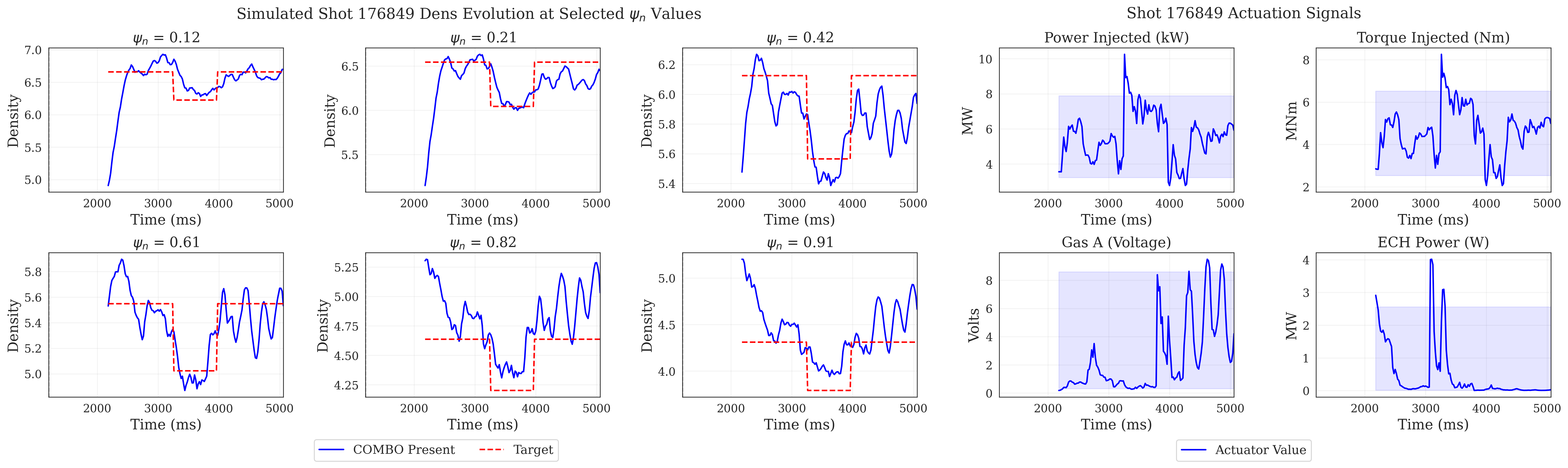}
    \caption{Lower density target}
  \end{subfigure}

  \begin{subfigure}{\textwidth}
    \centering
    \includegraphics[width=\linewidth]{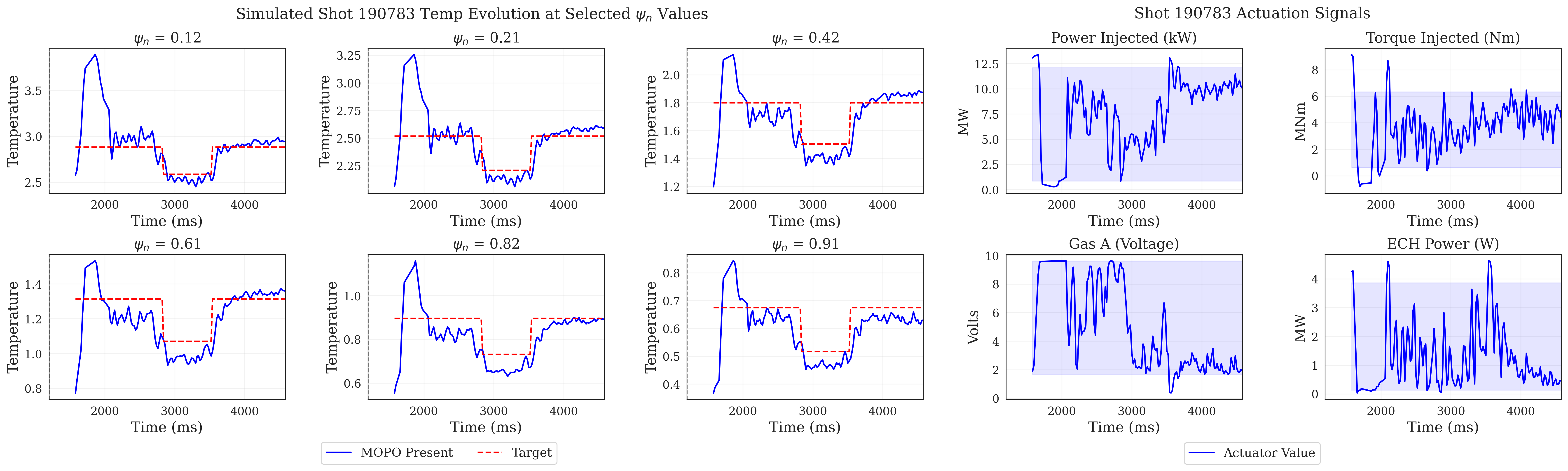}
    \caption{Lower temperature target}
  \end{subfigure}

   \begin{subfigure}{\textwidth}
    \centering
    \includegraphics[width=\linewidth]{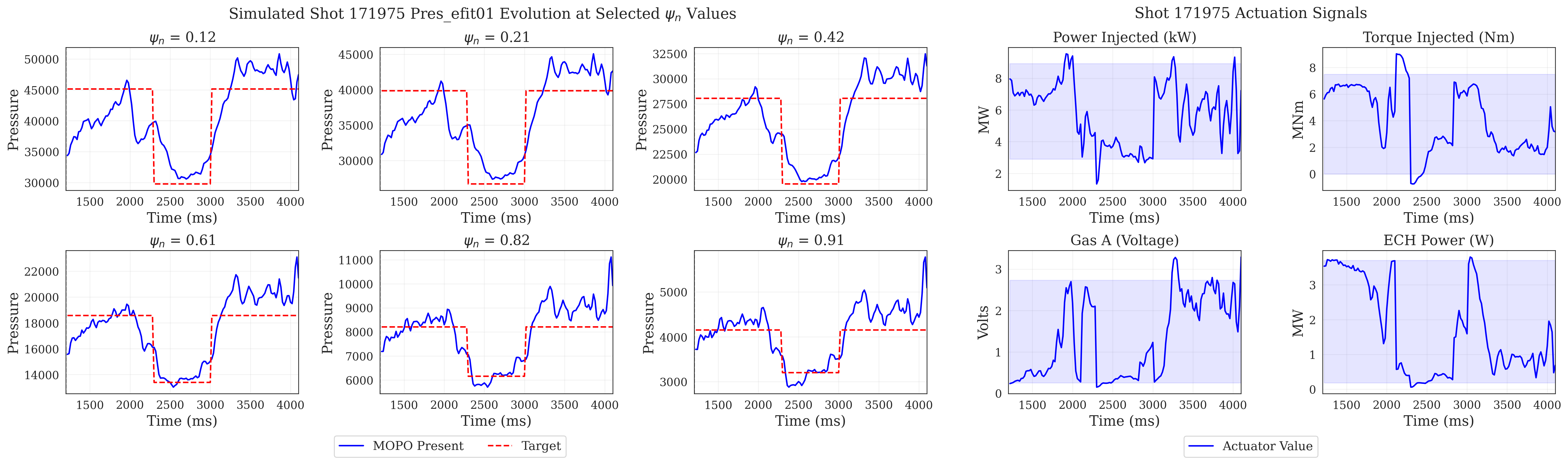}
    \caption{Lower pressure target}
  \end{subfigure}

  \caption{Simulated results for the lower-target cases of the remaining three tasks in the dynamics-model environment. From top to bottom, the panels show COMBO on density profile tracking (shot 176849), MOPO on temperature profile tracking (shot 190783), and MOPO on pressure profile tracking (shot 171975).}
  \label{fig:temp}
\end{figure}

\clearpage

\section{RPNN}
\label{app:rpnn}
\subsection{RPNN Training}
\label{app:rpnntrain}
The benchmark uses two learned dynamics models: a reference dynamics model trained on historical DIII-D experimental trajectories, and a separate dynamics model trained on the synthesized benchmark trajectories for model-based baselines. Both models use the same RPNN ensemble architecture described in Appendix~\ref{app:architecture}; they differ only in the data used for training. Each model predicts transitions over the full dynamics state--actuator space, which is larger than the policy observation and action space. The corresponding signal usage is summarized in Table~\ref{tab:plasma_signals}: variables marked as Actuators and States are used for dynamics modeling, while only a task-dependent subset of the state variables is exposed to the policy as observations.

We report held-out predictive fidelity using per-variable explained variance (EV). The reference dynamics model, trained on historical experimental data and used as the closed-loop evaluation environment, is summarized in Table~\ref{tab:dynamics-real}. The dynamics model trained on synthesized benchmark trajectories, used by model-based baselines for learned rollouts, is summarized in Table~\ref{tab:dynamics-synth}.

\begin{table}[ht]
    \centering
    \caption{Per-variable explained variance (EV) of the reference dynamics model trained on historical experimental data. Higher values indicate better agreement with held-out real trajectories.}
    \label{tab:dynamics-real}
    \small
    \setlength{\tabcolsep}{6pt}
    \begin{tabular}{l c l c}
        \toprule
        Variable & EV & Variable & EV \\
        \midrule
        $\beta_N$ & 0.8229 & itemp\_pca2 & 0.6044 \\
        dssdenest & 0.6658 & itemp\_pca3 & 0.4058 \\
        $l_i$ & 0.5374 & itemp\_pca4 & 0.2922 \\
        n1rms & 0.4520 & dens\_pca1 & 0.4584 \\
        $V_\text{loop}$ & 0.5187 & dens\_pca2 & 0.3938 \\
        $W_\text{MHD}$ & 0.8144 & dens\_pca3 & 0.3578 \\
        temp\_pca1 & 0.5810 & dens\_pca4 & 0.3938 \\
        temp\_pca2 & 0.4225 & rotation\_pca1 & 0.4473 \\
        temp\_pca3 & 0.3494 & rotation\_pca2 & 0.4016 \\
        temp\_pca4 & 0.3396 & rotation\_pca3 & 0.3004 \\
        itemp\_pca1 & 0.5878 & rotation\_pca4 & 0.3521 \\
        pres\_pca1 & 0.8853 & q\_pca1 & 0.3945 \\
        pres\_pca2 & 0.7107 & q\_pca2 & 0.4840 \\
        \bottomrule
    \end{tabular}
\end{table}

\begin{table}[ht]
    \centering
    \caption{Per-variable explained variance (EV) of the dynamics model trained on synthesized benchmark trajectories, evaluated on held-out data. Higher values indicate better predictive accuracy for model-based baselines.}
    \label{tab:dynamics-synth}
    \small
    \setlength{\tabcolsep}{6pt}
    \begin{tabular}{l c l c}
        \toprule
        Variable & EV & Variable & EV \\
        \midrule
        $\beta_N$ & 0.9647 & itemp\_pca2 & 0.9193 \\
        dssdenest & 0.8945 & itemp\_pca3 & 0.8776 \\
        $l_i$ & 0.9515 & itemp\_pca4 & 0.7848 \\
        n1rms & 0.8491 & dens\_pca1 & 0.8534 \\
        $V_\text{loop}$ & 0.8675 & dens\_pca2 & 0.7224 \\
        $W_\text{MHD}$ & 0.9678 & dens\_pca3 & 0.7117 \\
        temp\_pca1 & 0.8764 & dens\_pca4 & 0.8425 \\
        temp\_pca2 & 0.7221 & rotation\_pca1 & 0.9133 \\
        temp\_pca3 & 0.6276 & rotation\_pca2 & 0.8727 \\
        temp\_pca4 & 0.8003 & rotation\_pca3 & 0.8242 \\
        itemp\_pca1 & 0.9313 & rotation\_pca4 & 0.8100 \\
        pres\_pca1 & 0.9838 & q\_pca1 & 0.9365 \\
        pres\_pca2 & 0.9712 & q\_pca2 & 0.9428 \\
        \bottomrule
    \end{tabular}
\end{table}

\clearpage

\begin{table*}[ht]
\centering
\caption{
Plasma signals and how they are used as state and actuator variables for dynamics modeling. Policy observation and action space variables are also shown. Profile dimensionality depends on the task: rotation, density, and temperature use 4 PCA components, while pressure uses 2 PCA components.
}
\label{tab:plasma_signals}

\small
\setlength{\tabcolsep}{5pt}
\renewcommand{\arraystretch}{1.1}

\begin{tabular}{l p{0.21\textwidth} cccc}
\toprule
\textbf{Signal Group} & \textbf{Signals} & \textbf{Actuator} & \textbf{State} & \textbf{Action} & \textbf{Observation} \\
\midrule

\multirow{5}{*}{\textbf{Scalar States}}
& $\beta_N$ (Normalized Plasma Pressure) & \xmark & \cmark & \xmark & \xmark \\
& $l_i$ (Internal Inductance) & \xmark & \cmark & \xmark & \xmark \\
& Line Averaged Density & \xmark & \cmark & \xmark & \xmark \\
& Loop Voltage & \xmark & \cmark & \xmark & \xmark \\
& MHD Stored Energy & \xmark & \cmark & \xmark & \xmark \\
\midrule

\multirow{6}{*}{\textbf{Profile States}}
& Rotation & \xmark & \cmark & \xmark & \cmark\ (rot task) \\
& Density & \xmark & \cmark & \xmark & \cmark\ (dens task) \\
& Ion Temperature & \xmark & \cmark & \xmark & \xmark\  \\
& Electron Temperature & \xmark & \cmark & \xmark & \cmark\ (temp task) \\
& Pressure & \xmark & \cmark & \xmark & \cmark\ (pressure task) \\
& Safety Factor $q$ & \xmark & \cmark & \xmark & \xmark \\
\midrule

\multirow{5}{*}{\textbf{Shape Variables}}
& Elongation & \multirow{5}{*}{\cmark} & \multirow{5}{*}{\xmark} & \multirow{5}{*}{\xmark} & \multirow{5}{*}{\xmark} \\
& Upper Triangularity & & & & \\
& Bottom Triangularity & & & & \\
& $a_{\text{minor}}$ & & & & \\
& Radial and vertical positions of magnetic axis & & & & \\
\midrule

\textbf{Neutral Beam Variables}
& Power Injected & \cmark & \xmark & \cmark & \xmark \\
& Torque Injected & \cmark & \xmark & \cmark & \xmark \\
\midrule

\textbf{Gas Puffing}
& GasA voltage & \cmark & \xmark & \cmark & \xmark \\
\midrule

\textbf{Electron Cyclotron Heating}
& ECH Total Power & \cmark & \xmark & \cmark & \xmark \\
\midrule

\textbf{Other Actuators}
& Current Target, Toroidal Field & \cmark & \xmark & \xmark & \xmark \\
\midrule

\multirow{4}{*}{\textbf{Targets}} 
& Rotation Target $(t)$,       &            &            &            &            \\
& Rotation Target $(t+10)$,    & \xmark     & \xmark     & \xmark     & \cmark     \\
& Error Terms $(t)$          &            &            &            &            \\
& Error Terms $(t+10)$          &            &            &            &            \\
\midrule

\textbf{Total Dimensions} & & \textbf{12D} & \textbf{25D} & \textbf{4D} & \textbf{20D/10D(pres task)} \\
\bottomrule
\end{tabular}
\end{table*}
\subsection{Network Architecture and Training Details}
\label{app:architecture}
\paragraph{Network Architecture}
\begin{itemize}
    \item \textbf{Encoder:}
    \begin{itemize}
        \item Fully Connected (FC) layer: $\texttt{input\_dim} \times 512$
        \item FC layer: $512 \times 512$
    \end{itemize}

    \item \textbf{Memory Unit:}
    \begin{itemize}
        \item Gated Recurrent Unit (GRU) block: $512 \times 256$
    \end{itemize}

    \item \textbf{Decoder} (with residual connections between FC layers):
    \begin{itemize}
        \item FC layer: $256 \times 512$
        \item FC layers: $512 \times 512$ (repeated 8 times)
        \item FC layer: $512 \times 128$
    \end{itemize}

    \item \textbf{Output Heads:}
    \begin{itemize}
        \item Mean head: $128 \times \texttt{output\_dim}$
        \item Log-variance head: $128 \times \texttt{output\_dim}$
    \end{itemize}
\end{itemize}

\section{Implementation Details}
\label{app:implementationdetails}
\subsection{Hyperparameter}
We separate hyperparameter selection from final evaluation. For each task and algorithm, hyperparameters were selected using the validation split, while the held-out test split was used only for the final closed-loop evaluation reported in the main results. This protocol avoids tuning directly on the test shots and provides a less biased estimate of generalization performance. Because the four profile-control tasks differ in their response channels and long-horizon dynamics, we allow a small set of task-dependent hyperparameters, but keep all other settings shared across tasks or fixed by the released configuration files.

Table~\ref{tab:common_hparams} summarizes the common training, evaluation, network, and model-rollout settings used across the benchmark. These settings define the default experimental budget and architecture choices. Method-specific deviations are limited to the hyperparameters listed in Table~\ref{tab:benchmark-hparams}.

For hyperparameter tuning, we used an Optuna-based tuner implemented in the released codebase. The tuner reads each algorithm's search space from the configuration file, launches training trials with sampled hyperparameters, and maximizes the validation reward extracted from the training logs. The default sampler is Optuna's TPE sampler, with support for grid search when specified in the configuration. Trials can be run in parallel across multiple GPUs, and the tuner records the best trial, selected parameters, log directory, and detailed trial results.

Table~\ref{tab:task_dependent_hparams} reports the task-dependent hyperparameters selected by this validation procedure. These values were fixed before evaluating on the test split. Parameters not shown in the table are either shared across all four tasks, listed in Table~\ref{tab:common_hparams}, or specified in the released per-method configuration files.

\begin{table}[ht]
\centering
\small
\setlength{\tabcolsep}{5pt}
\renewcommand{\arraystretch}{1.15}
\caption{Common experimental settings used in the benchmark. We report the main shared settings here and provide complete per-method configurations in the released codebase.}
\label{tab:common_hparams}

\begin{tabularx}{\textwidth}{
>{\RaggedRight\arraybackslash}p{2.7cm}
>{\RaggedRight\arraybackslash}p{3.5cm}
>{\RaggedRight\arraybackslash}p{3.0cm}
>{\RaggedRight\arraybackslash}X
}
\toprule
\textbf{Category} & \textbf{Setting} & \textbf{Value} & \textbf{Notes} \\
\midrule

\multicolumn{4}{l}{\textbf{Training}} \\
\midrule
Optimization budget 
& Gradient updates 
& \texttt{1000} epochs, \texttt{1000} steps/epoch 
& Used by most offline RL baselines. \\

Mini-batch size 
& Default batch size 
& \texttt{256} 
& GCIL and PPO use method-specific batch sizes. \\

Discount factor 
& \texttt{gamma} 
& \texttt{0.99} 
& Used by most offline RL baselines; PPO use separate settings. \\

Target update 
& \texttt{tau} 
& \texttt{0.005} 
& Used by actor-critic offline RL baselines. \\

\midrule
\multicolumn{4}{l}{\textbf{Evaluation}} \\
\midrule
Test set 
& Held-out shots 
& \texttt{300} shots 
& Fixed test split shared across all tasks. \\

Randomization 
& Seeds per shot 
& \texttt{10} 
& Used for closed-loop policy evaluation. \\

Metric 
& Tracking error 
& RMSE $\downarrow$ 
& Reported with standard error across rollout instances. \\

Evaluation episodes 
& Default value 
& \texttt{5} 
& PPO uses \texttt{3} evaluation episodes. \\

\midrule
\multicolumn{4}{l}{\textbf{Policy and value networks}} \\
\midrule
Default MLP 
& Hidden dimensions 
& \texttt{[256,256]} 
& Used by several actor-critic and model-based baselines. \\

Larger MLP 
& Hidden dimensions 
& \texttt{[256,256,256]} 
& Used by COMBO, CQL, EDAC, GCIL, and MPPI. \\

PPO architecture 
& Policy/value networks 
& \texttt{[256,256]} 
& PPO uses separate policy and value networks. \\

\midrule
\multicolumn{4}{l}{\textbf{Model-based rollouts}} \\
\midrule
Rollout batch size 
& \texttt{rollout\_batch\_size} 
& \texttt{50000} 
& Used by model-based offline RL baselines. \\

Model retention 
& \texttt{model\_retain\_epochs} 
& \texttt{5} 
& Controls the retained model-generated replay buffer. \\

Rollout schedule 
& \texttt{rollout\_freq} 
& \texttt{1000} 
& Default rollout frequency for model-based methods. \\

Dynamics ensemble 
& Number of models 
& \texttt{25} 
& Used for learned dynamics-model rollouts and closed-loop evaluation. \\

\bottomrule
\end{tabularx}
\end{table}
\begin{table}
    \centering
    \caption{Algorithms evaluated in the benchmark and the hyperparameters tuned for each method. BAMCTS is implemented as \texttt{bambrl} in the codebase.}
    \label{tab:benchmark-hparams}
    \footnotesize
    \setlength{\tabcolsep}{3.5pt}
    \renewcommand{\arraystretch}{1.05}
    \begin{center}
    \begin{minipage}{0.6\textwidth}
    \centering
    \begin{tabularx}{\linewidth}{@{}llX@{}}
        \toprule
        \textbf{Category} & \textbf{Algorithm} & \textbf{Tuned hyperparameters} \\
        \midrule
        Imitation Learning
        & GCIL & batch size \\
        \midrule
        \multirow{5}{*}{\shortstack[l]{Model-free\\offline RL}}
        & TD3BC & $\alpha$ \\
        & CQL   & CQL weight, temperature \\
        & IQL   & expectile, temperature \\
        & EDAC  & number of critics, $\eta$ \\
        & MCQ   & $\lambda$, sampled actions \\
        \midrule
        \multirow{6}{*}{\shortstack[l]{Model-based\\offline RL}}
        & PPO   & clip range \\
        & COMBO  & rollout length, CQL weight \\
        & MOPO   & rollout length, penalty coef. \\
        & MOBILE & rollout length, penalty coef. \\
        & RAMBO  & rollout length, adv.\ weight \\
        & \multirow{2}{*}{BAMCTS} & rollout length, penalty coef., \texttt{use\_ba}, \\
        &        & \texttt{search\_alpha} $\in \{0.5, 0.8\}$ \\
        \bottomrule
    \end{tabularx}
    \end{minipage}
    \end{center}
\end{table}
\begin{table}[!htbp]
\centering
\small
\setlength{\tabcolsep}{6pt}
\renewcommand{\arraystretch}{1.12}
\caption{Task-dependent hyperparameters. All other hyperparameters are shared across the four benchmark tasks or specified in the released configuration files.}
\label{tab:task_dependent_hparams}

\begin{tabular}{llcccc}
\toprule
\textbf{Method} & \textbf{Hyperparameter} 
& \textbf{Rotation} & \textbf{Density} & \textbf{Temperature} & \textbf{Pressure} \\
\midrule
GCIL                   & \texttt{batch\_size}     & 64 & 512 & 64 & 256 \\
\midrule
\multirow{1}{*}{PPO}   & \texttt{clip\_range}     & 0.148 & \textemdash & \textemdash & \textemdash \\
                       
\midrule
TD3BC                  & \texttt{alpha}           & 1.5 & 0.15 & 2.0 & 5.0 \\
\midrule
\multirow{2}{*}{CQL}   & \texttt{cql\_weight}     & 2.0 & 2.0 & 5.0 & 0.1 \\
                       & \texttt{temperature}     & 1.0 & 2.0 & 5.0 & 0.1 \\
\midrule
\multirow{2}{*}{IQL}   & \texttt{expectile}       & 0.67 & 0.69 & 0.52 & 0.55 \\
                       & \texttt{temperature}     & 0.21 & 4.10 & 0.67 & 0.82 \\
\midrule
\multirow{2}{*}{EDAC}  & \texttt{num\_critics}    & 10  & 10  & 50  & 20  \\
                       & \texttt{eta}             & 0.1 & 2.0 & 2.0 & 0.5 \\
\midrule
\multirow{2}{*}{MCQ}   & \texttt{lmbda}           & 0.771114 & 0.750130 & 0.860632 & 0.711199 \\
                       & \texttt{num\_sampled\_actions} & 20 & 10 & 10 & 20 \\
\midrule
\multirow{2}{*}{COMBO} & \texttt{cql\_weight}     & 1   & 3   & 5   & 3.0 \\
                       & \texttt{rollout\_length} & 10  & 7   & 10  & 5   \\
\midrule
\multirow{2}{*}{MOPO}  & \texttt{rollout\_length} & 7   & 5   & 7   & 7   \\
                       & \texttt{penalty\_coef}   & 0.5 & 5.0 & 1   & 1   \\
\midrule
\multirow{2}{*}{MOBILE} & \texttt{rollout\_length} & 1 & \textemdash & \textemdash & \textemdash \\
                       & \texttt{penalty\_coef}   & 1.5 & \textemdash & \textemdash & 0.5 \\
\midrule
\multirow{2}{*}{RAMBO} & \texttt{rollout\_length} & 5 & 2 & 5 & 4 \\
                       & \texttt{adv\_weight}     & 0.000001 & 0.000408 & 0.000095 & 0.000023 \\
\midrule
\multirow{4}{*}{BAMCTS} & \texttt{rollout\_length} & 5 & 2 & 1 & 5 \\
                       & \texttt{penalty\_coef}   & 1.72 & 0.92 & 1.29 & 1.64 \\
                       & \texttt{use\_ba}         & \texttt{False} & \texttt{False} & \texttt{True} & \texttt{False} \\
                       & \texttt{search\_alpha}   & 0.50 & 0.50 & 0.50 & 0.50 \\
\bottomrule
\end{tabular}
\end{table}

\clearpage

\subsection{Experiments compute resources}
\label{app:computeresources}
All experiments were conducted on an Ubuntu 22.04 Linux server equipped with two Intel Xeon Platinum 8457C processors, providing 96 physical CPU cores (192 threads) in total, 503 GiB RAM, and 10 NVIDIA GeForce RTX 4090 GPUs. Each GPU provides approximately 46-49 GiB of memory. Unless otherwise specified, each training run used a single GPU. Training time varies across algorithm classes. Model-based methods generally require longer training due to learned-model rollouts, with representative runs ranging from several to about 30 GPU-hours (MCQ: 3.7h, MOPO: 11h, BAMCTS: 12.0h, MOBILE: 17h, RAMBO: 24.5h, COMBO: 1.2 days), whereas model-free and imitation-learning methods are comparatively faster, typically completing within minutes to 18 GPU-hours (GCIL: 43 min, IQL: 1.8h, CQL: 18h, EDAC: 8h, TD3-BC: 6h).

\section{Broader Impacts}
\label{app:broaderimpacts}
RL4F is an open-source benchmark for offline reinforcement learning in nuclear-fusion plasma profile control, enabling researchers to develop data-driven control algorithms and compare them under standardized tasks, datasets, dynamics-model environments, and evaluation protocols. By supporting reproducible offline evaluation, RL4F can help reduce the need for costly and safety-critical trial-and-error on real tokamak devices during early-stage controller development. Progress on this problem may contribute to more effective regulation of plasma profiles related to confinement, fueling, heating, and overall plasma performance, which is an important step toward stable and efficient fusion operation. More broadly, the benchmark provides a challenging real-world testbed for offline RL research, with nonlinear dynamics, long-horizon control, partial observability, and limited data coverage. As with any simulation-based benchmark in a safety-critical domain, performance in RL4F should be interpreted as evidence from a controlled offline evaluation setting rather than as a substitute for real-device validation; any potential deployment on physical tokamaks would require additional safety constraints, expert review, and device-specific testing.

\end{document}